\RequirePackage{fix-cm}
\documentclass{svjour3}                     
\smartqed  

\usepackage{amsmath}
\usepackage{amssymb}
\usepackage{booktabs}
\usepackage{cleveref}
\usepackage{graphicx}
\usepackage{multirow}
\usepackage{subcaption}
\usepackage{threeparttable}
\usepackage{url}
\usepackage{wasysym}

\crefname{figure}{Fig.}{Figs.}
\crefname{table}{Table.}{Tables.}

\newcommand{\dash}[0]{\multicolumn{1}{c}{-}}

\journalname{Multimedia Tools and Applications}

\begin{document}

\title{Object Detection for Comics using Manga109 Annotations}

\author{
  Toru Ogawa \and
  Atsushi Otsubo \and
  Rei Narita \and
  Yusuke Matsui \and
  Toshihiko Yamasaki \and
  Kiyoharu Aizawa
}


\institute{
  T. Ogawa \at The University of Tokyo
  \and
  A. Otsubo \at The University of Tokyo
  \and
  R. Narita \at The University of Tokyo
  \and
  Y. Matsui \at National Institute of Informatics, Japan
  \and
  T. Yamasaki \at The University of Tokyo
  \and
  K. Aizawa \at The University of Tokyo
}

\date{Received: date / Accepted: date}

\maketitle

\begin{abstract}
With the growth of digitized comics, image understanding techniques are becoming important.
In this paper, we focus on object detection, which is a fundamental task of image understanding.
Although convolutional neural networks (CNN)-based methods archived good performance in object detection for naturalistic images,
there are two problems in applying these methods to the comic object detection task.
First, there is no large-scale annotated comics dataset.
The CNN-based methods require large-scale annotations for training.
Secondly, the objects in comics are highly overlapped compared to naturalistic images.
This overlap causes the assignment problem in the existing CNN-based methods.
To solve these problems, we proposed a new annotation dataset and a new CNN model.
We annotated an existing image dataset of comics and created the largest annotation dataset, named Manga109-annotations.
For the assignment problem, we proposed a new CNN-based detector, SSD300-fork.
We compared SSD300-fork with other detection methods using Manga109-annotations
and confirmed that our model outperformed them based on the mAP score.

\end{abstract}

\section{Introduction}
\label{sec:intro}
Comics, defined as a form of juxtaposed sequences of panels of images,
are a popular medium, not only for the entertainment, but also as an important part of the cultural heritage.
With the rapid growth of mobile technologies,
digitized comics are becoming important for the distribution of them over mobile devices, such as the Amazon Kindle and iPad.
To make use of such digitized comics for further applications,
understanding the elements in comics would play an essential role.

\begin{figure}[!t]
  \centering
  \begin{minipage}{0.4\hsize}
    \frame{\includegraphics[width=1\hsize]{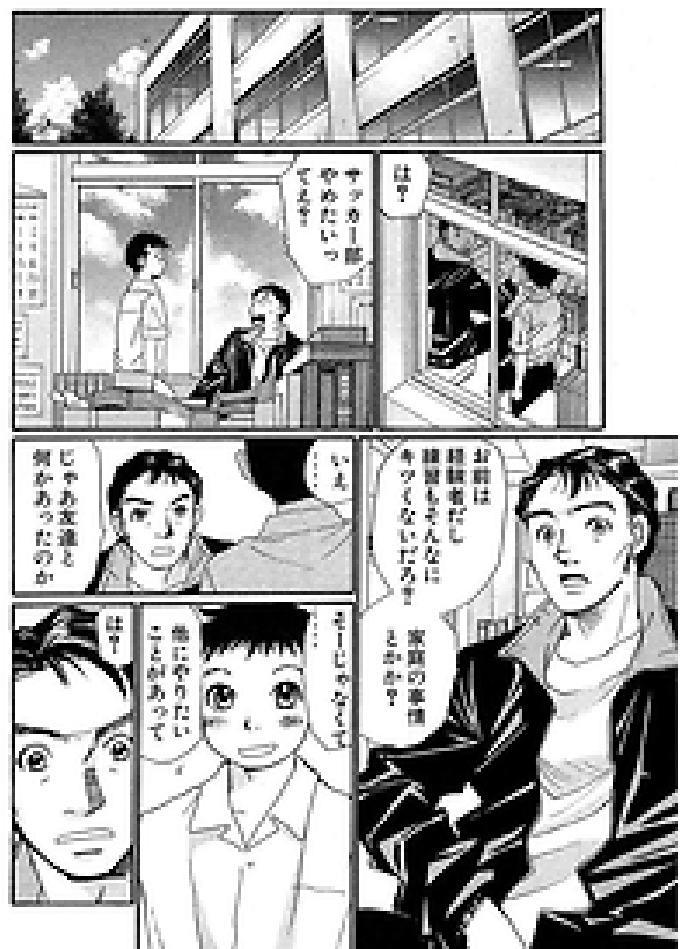}}
    \subcaption{Input \footnotemark[1]}
    \label{fig:intro:overview:input}
  \end{minipage}
  {\Huge\pointer}
  \begin{minipage}{0.4\hsize}
    \frame{\includegraphics[width=1\hsize]{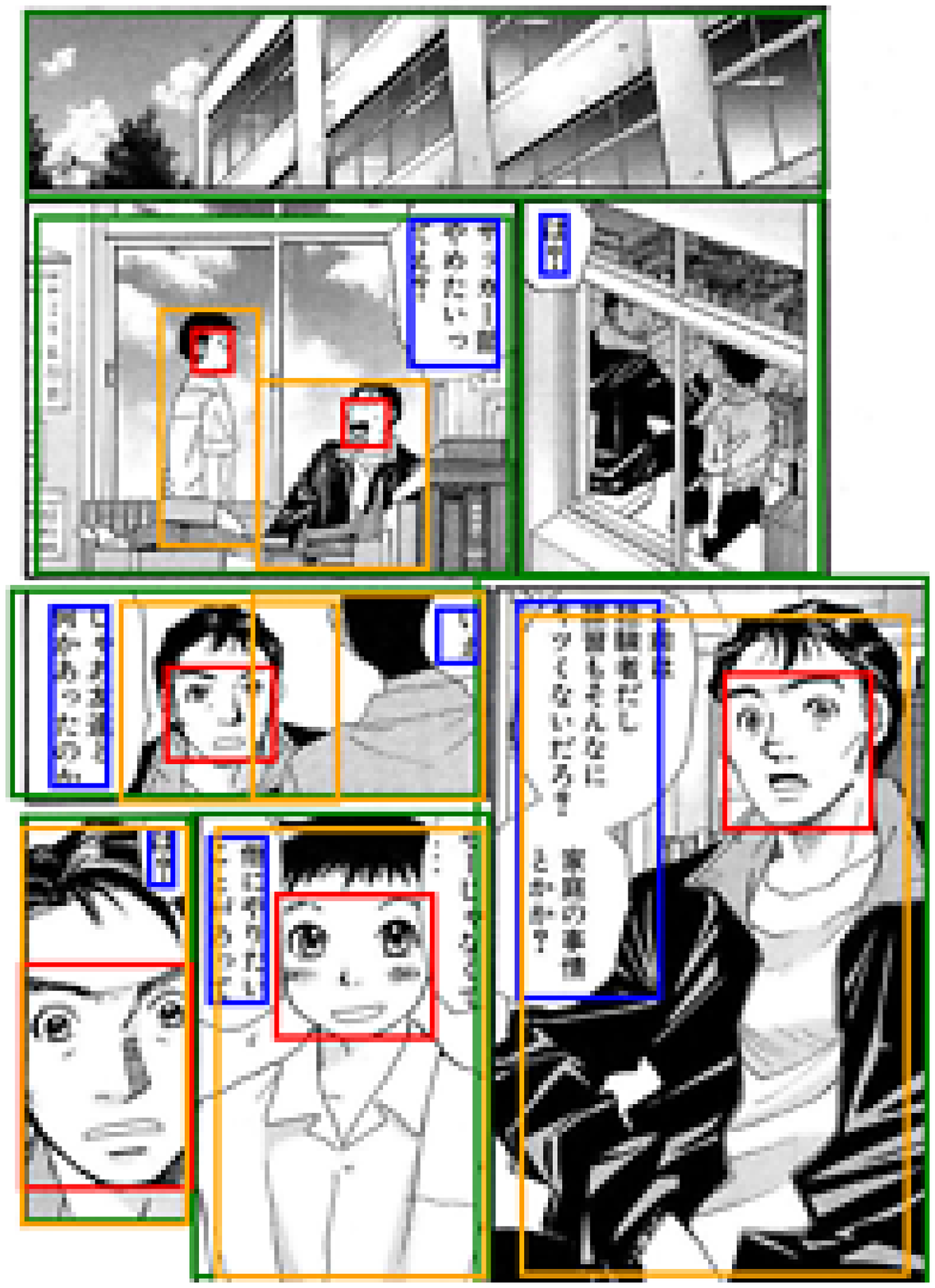}}
    \subcaption{Output \footnotemark[1]}
    \label{fig:intro:overview:output}
  \end{minipage}
  \caption{Comic object detection.
    The objective is, given a comic page (\subref{fig:intro:overview:input}),
    to predict the location and category of each component (\subref{fig:intro:overview:output}).}
  \label{fig:intro:overview}
\end{figure}

In particular, we focus on the object detection task over comics (\cref{fig:intro:overview}).
Given an image (page) of comics, our objective is to automatically detect
visual elements such as frames, faces, and texts.
Such automatic detection is a fundamental building block of analyzing the contents of comics,
and required for a number of applications including retargeting and automatic translation.
Retargeting is a task whereby the contents of comics are reorganized
so that they can be displayed them in other media that have different aspect ratios.
Detecting frames is an essential factor to achieve retargeting.
Automatic translation for comics is widely in demand.
It can be implemented using a combination of text detection, optical character recognition (OCR), and machine translation.
\footnotetext[1]{``YamatoNoHane'' \textcopyright Saki Kaori}

\begin{figure}[!t]
  \centering
  \frame{\includegraphics[width=0.6\hsize]{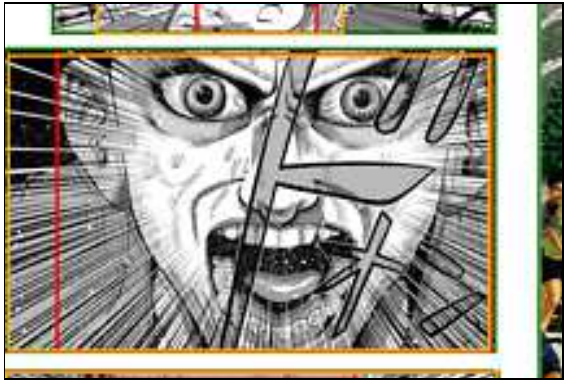}}
  \caption{Highly overlapped objects in comics \protect\footnotemark[2].
    \textit{frame} (green), \textit{face} (red), and \textit{body} (orange) are highly overlapped.}
  \label{fig:relatedwork:dense}
\end{figure}

Although the object detection for naturalistic images have been widely researched~\cite{ren2015faster,redmon2016yolo9000,liu2016ssd,fu2017dssd},
there are two fundamental difficulties in successfully applying such methods to comics.
First, there are no large-scale annotations for comics.
To achieve sufficient performance, detailed and large-scale annotations such as
bounding boxes of characters would be required,
but such datasets do not yet exist.
Secondly, even though a detector can be trained, an \textit{assignment problem} emerges.
\Cref{fig:relatedwork:dense} illustrates an example, where the three elements (\textit{frame}, \textit{face}, and \textit{body}) are highly overlapped.
Compared to the naturalistic images, such overlap often occurs
because of the construction of comics, i.e., semantically important elements tend to fill the entire area of the frame.
In such cases, state-of-the-art anchor-based detectors cannot work well
because some overlapped objects might not be assigned to any anchor boxes.
\footnotetext[2]{``KyokugenCyclone'' \textcopyright Takanami Shin}

To solve the two problems above, we developed a large-scale annotation dataset and an object detector suited for comics.
The contribution of this paper is as follows:
\begin{itemize}
\item We created \textit{Manga109-annotations}, detailed and large-scale annotations for the Manga109 image dataset~\cite{matsui2016sketch}.
  Manga109 is a large scale comic dataset containing 21,142 images.
  We assigned 527,685 annotations over the whole dataset with the help of 72 workers in six months.
  To improve the quality of annotations, we conducted additional verification and refinement with 26 workers in two months.
\item Based on the SSD architecture~\cite{liu2016ssd},
  we designed \textit{SSD300-fork}, an object detector tailored for the highly overlapped situation.
  By replicating the anchor set, the proposed SSD300-fork successfully detects overlapped objects.
  Experimental results show that SSD300-fork outperforms the state-of-the-art detection methods at the mAP score.
\end{itemize}

\section{Related work}
\label{sec:related}

\subsection{Comic datasets}
\label{sec:related:dataset}
We first review currently available comics datasets (\cref{table:dataset}).
Due to the copyright issue,
the number of datasets of comics that can be used for academic research is limited.

Gu{\'e}rin et al. published eBDtheque~\cite{eBDtheque2013}.
This dataset contains 100 images from French, American, and Japanese comics.
In addition, it provides manual annotations including the position of frames and speech balloons.
We could use these annotations to evaluate object detection tasks
because they contain (manually annotated) location information.
However, it is difficult to use this dataset for machine learning methods
because the number of images is too small.

Mohit et al. published COMICS~\cite{IyyerComics2016}, which is the largest-scale comic dataset for research usage.
This dataset contains 3,948 volumes of American comics.
They also provided bounding box annotations of frame and text.
However, these annotations are semi-automatically created, and
would be suitable for neither training nor evaluation
(500 pages are manually annotated,
but the annotations attached to the rest pages are the result of the detection of Faster~R-CNN~\cite{ren2015faster} trained by the 500 pages).

Matsui et al. published Manga109~\cite{matsui2016sketch},
which contains 109 volumes of Japanese comics (manga) drawn by 104 authors and 21,142 images.
These comics have two desired characteristics: (a) high quality and (b) large diversity.
(a) High quality: all volumes were drawn by professional authors and commercially published.
(b) Large diversity: The years of publication are distributed from the 1970s to the 2010s, and 11 genres (sports, romantic comedy, etc.) are covered.
Although Manga109 has these advantages, no annotations are provided.
We annotate this dataset for the object detection task,
where the details of the annotation are explained in \cref{sec:dataset}.

\begin{table*}[!t]
  \caption{Comic datasets and Manga109-annotations}
  \label{table:dataset}
  \centering
  \begin{threeparttable}
    {\tabcolsep=1mm \begin{tabular}{@{}llrrrrrr@{}}
      \toprule
      &&&&\multicolumn{4}{c}{Annotations} \\ \cmidrule(l){5-8}
      Dataset&&\#volume&\#page&\#frame&\#text&\#face&\#body \\ \midrule
      eBDtheque~\cite{eBDtheque2013}&&25&100&850&1,092&\dash&1,550 \\
      COMICS~\cite{IyyerComics2016}&&3,948&198,657&\small{(1,229,664)} \tnote{$\dagger$}&\small{(2,498,657)} \tnote{$\dagger$}&\dash&\dash \\
      Manga109~\cite{matsui2016sketch}&&109&21,142&\dash&\dash&\dash&\dash \\ \midrule
      Manga109-annotations&total&109&10,130 \tnote{$\ddagger$}&103,900&147,918&118,715&157,152 \\
      &train&99&9,250 \tnote{$\ddagger$}&94,746&135,376&108,353&144,448 \\
      &test&10&880 \tnote{$\ddagger$}&9,154&12,542&10,362&12,704 \\
      \bottomrule
    \end{tabular}}
    \begin{tablenotes}[para]
    \item[$\dagger$] Pseudo annotation (semi-automatically annotated).
    \item[$\ddagger$] Double-sided page.
    \end{tablenotes}
  \end{threeparttable}
\end{table*}

\subsection{Comic researches based on object detection}
Chu et al. \cite{chu2014line,chu2016manga} proposed a method to compute a feature describing the styles of comics.
They used the positions and shapes of frames, speech balloons, and characters for computing this feature.
Rigaud et al. \cite{rigaud2015speech} proposed a method to predict the speaker of a given speech balloon.
This method requires character and speech balloon detection as the first step.
Le et al. \cite{le2015content} proposed a comic image retrieval method.
This method uses the positions of frames as features of pages.
Aramaki et al. \cite{aramaki2016text} proposed a text recognition method for comics.
They conducted text detection using low-level features.
Our large-scale comics dataset will be useful for these studies as training/evaluating data.

\subsection{Object detection for naturalistic images}
\label{sec:related:natural-detection}

Object detection for naturalistic images has been developed in many studies.
In this area, convolutional neural networks (CNNs) archived high performance
thanks to large scale datasets, such as PASCAL VOC~\cite{Everingham15} and MS COCO~\cite{lin2014microsoft}.
The most basic approach of CNN-based methods is R-CNN~\cite{girshick2014rich}.
It treats object detection as classification of regions proposed by selective search~\cite{uijlings2013selective}.
Fast~R-CNN~\cite{girshick2015fast} and Faster~R-CNN~\cite{ren2015faster} are the successors of R-CNN.
Fast~R-CNN introduced RoI Pooling to reduce the computation cost,
and Faster~R-CNN replaced the selective search with the region proposal network (RPN) for end-to-end training.

As a different approach from R-CNN, there exist anchor-based methods,
including YOLO~\cite{redmon2016you}, YOLOv2~\cite{redmon2016yolo9000}, SSD~\cite{liu2016ssd}, DSSD~\cite{fu2017dssd}, FPN~\cite{lin2017feature}, and RetinaNet~\cite{lin2017focal}.
These methods use anchor boxes instead of regions generated by selective search or RPN.
An anchor box is a rectangle region whose location and shape are fixed.
An anchor-based method has a set of regularly placed anchor boxes (we call this set \textit{anchor set}).
Each anchor box is trained to return the location and category of the nearest object.

\section{Dataset}
\label{sec:dataset}

In this section, we introduce \textit{Manga109-annotations},
a new annotation dataset based on Manga109~\cite{matsui2016sketch}.
As the original Manga109 does not contain annotations,
we manually annotated regions over the dataset.
We found an important region from an input image and assigned a category label to the region.
With our Manga109-annotations, one can train and evaluate comic object detector easily.
Manga109-annotations contains (1) bounding box annotations for the object detection task and
(2) additional annotations including character names and contents of texts, for further future studies.

\subsection{Bounding box annotations}
\begin{figure}[!t]
  \centering
  \includegraphics[width=0.7\hsize]{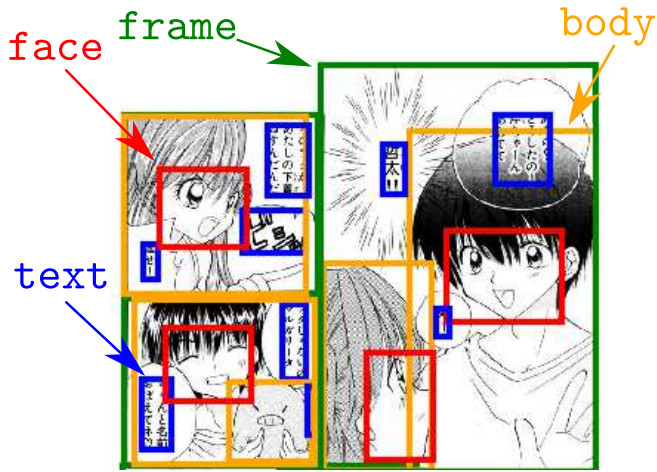}
  \caption{Four object categories \protect\footnotemark[3].
    These elements are fundamental and play important roles in comics.}
  \label{fig:dataset:annotation}
\end{figure}

Let us formally define the term, bounding box annotation.
A bounding box annotation consists of a bounding box and a category label.
A bounding box consists of four values, $(x_{min}, y_{min}, x_{max}, y_{max})$,
and represents a rectangular region in an image.
A category label indicates one of the following categories: \textit{frame}, \textit{text}, \textit{face}, and \textit{body}.
We selected these four categories
because these elements play important roles in comics as explained in the paragraphs that follow.
\Cref{fig:dataset:annotation} illustrates an example of annotations.
A bounding box is illustrated as a rectangle, and its color indicates the category label.
In the annotation process, workers selected each region that belongs to one of these categories with a bounding box
and assigned a category label to the region.

\textit{frame} is a region of page that describes a scene.
Typically, a page in comics is composed of a set of frames.
Visual elements (characters, background, texts, etc.) are usually drawn inside a frame.
Readers follow these frames in a certain order to understand the story.
A frame is usually described by a rectangle with a black border, but the shape can be of any polygon.
As a frame is a basic unit of image representation in comics, detecting frames is important.
Although several frame detection methods have been proposed to date~\cite{arai2011method,rigaud2013robust,rigaud2015knowledge},
their performances can not be compared because of the lack of publicly available groundtruth annotations.
Our Manga109-annotations provide such groundtruth, by which several frame detection methods can be evaluated and compared.

\textit{text} is an element that contains a character's speech, monologue, or narration.
Most of texts are placed in bounded regions filled by white color, i.e., they are overlaid on characters or background.
We call these regions \textit{speech balloons}.
Some of texts are placed directly without speech balloons.

\textit{face} and \textit{body} are the face and the whole body of a character, respectively.
We defined the region of a face as a rectangle including eyebrows, cheek, and chin.
The region of a body is defined as a rectangle including all body parts, such as the head, hair, arms, and legs.
As the face is also a part of body,
we can say a face is always included in the body region.
Sometimes, a body contains more than one face.
Some comics use animals, such as dogs and cats, as main characters.
In these cases, we treated them as we did human characters.
As faces and bodies are usually placed inside a frame,
it often happens that only a partial area can be observed.
For example, only an upper body of a boy is drawn in \cref{fig:dataset:annotation}.
In this case, the bounding box is bounded by the frame and the occluded part is not annotated.
\footnotetext[3]{``BakuretsuKungFuGirl'' \textcopyright Ueda Miki}

\subsection{Additional annotations}
For future studies, we assigned two types of additional information: character names and contents of texts.
Character names are assigned to all \textit{face} and \textit{body}.
2,979 unique character names are registered.
Each \textit{text} has its content as a unicode string.
The workers were asked to input the content of each text manually.
As a result of this annotation, we got 2,037,046 letters (5.7~MiB) text data.
Fully-manually collected text data with bounding boxes over 109 volumes are large.
To the best of our knowledge, a large dataset comparable with ous has not been developed yet.
These text data can be quite useful for a number of applications,
including automatic translation, text recognition, speaker detection, and multimodal learning.

\subsection{Annotation process}
\begin{figure}[!t]
  \centering
  \begin{minipage}{0.45\hsize}
    \frame{\includegraphics[width=1\hsize]{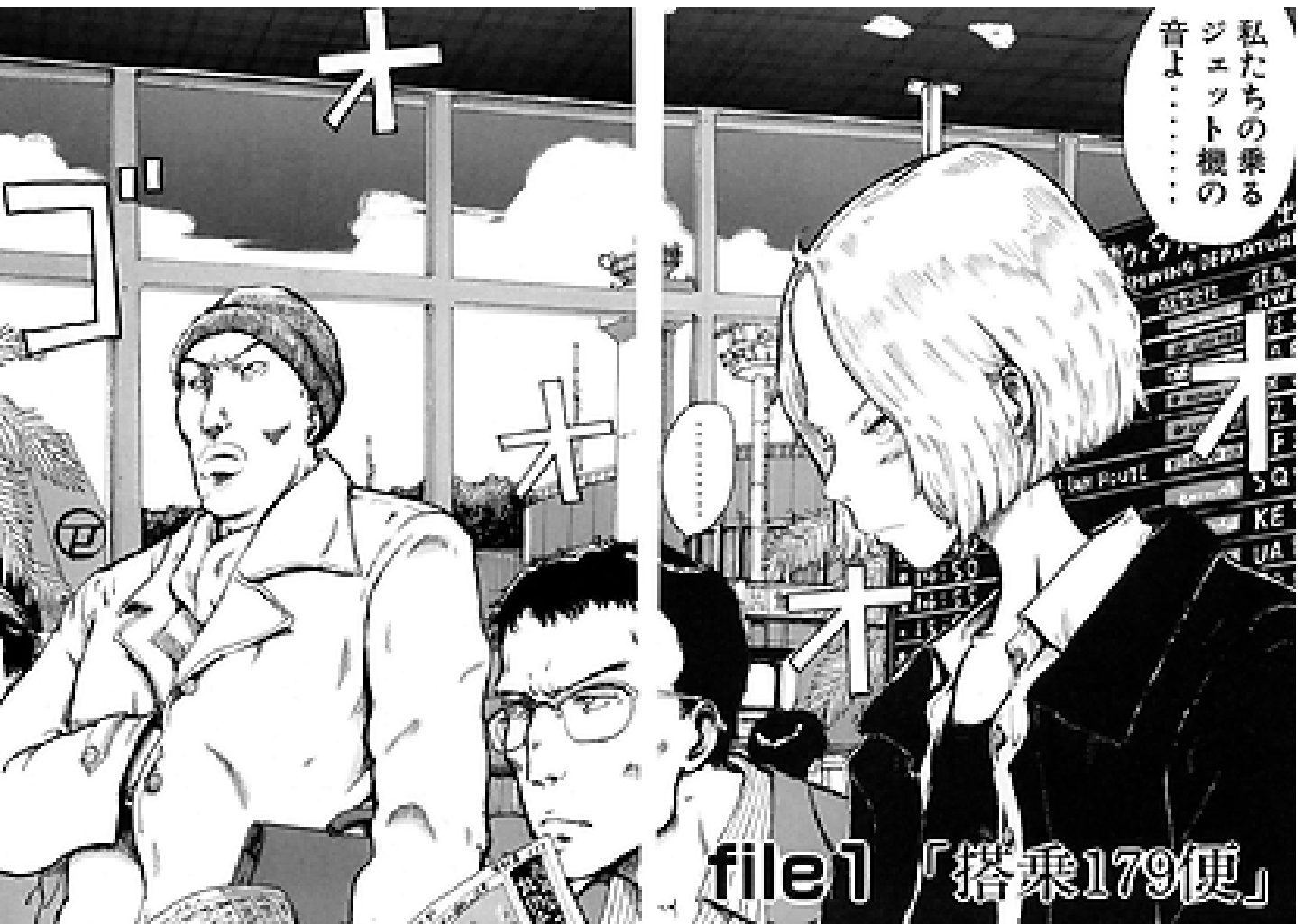}}
    \subcaption{\footnotemark[4]}
    \label{fig:dataset:double-sided:a}
  \end{minipage}
  \hfil
  \begin{minipage}{0.45\hsize}
    \frame{\includegraphics[width=1\hsize]{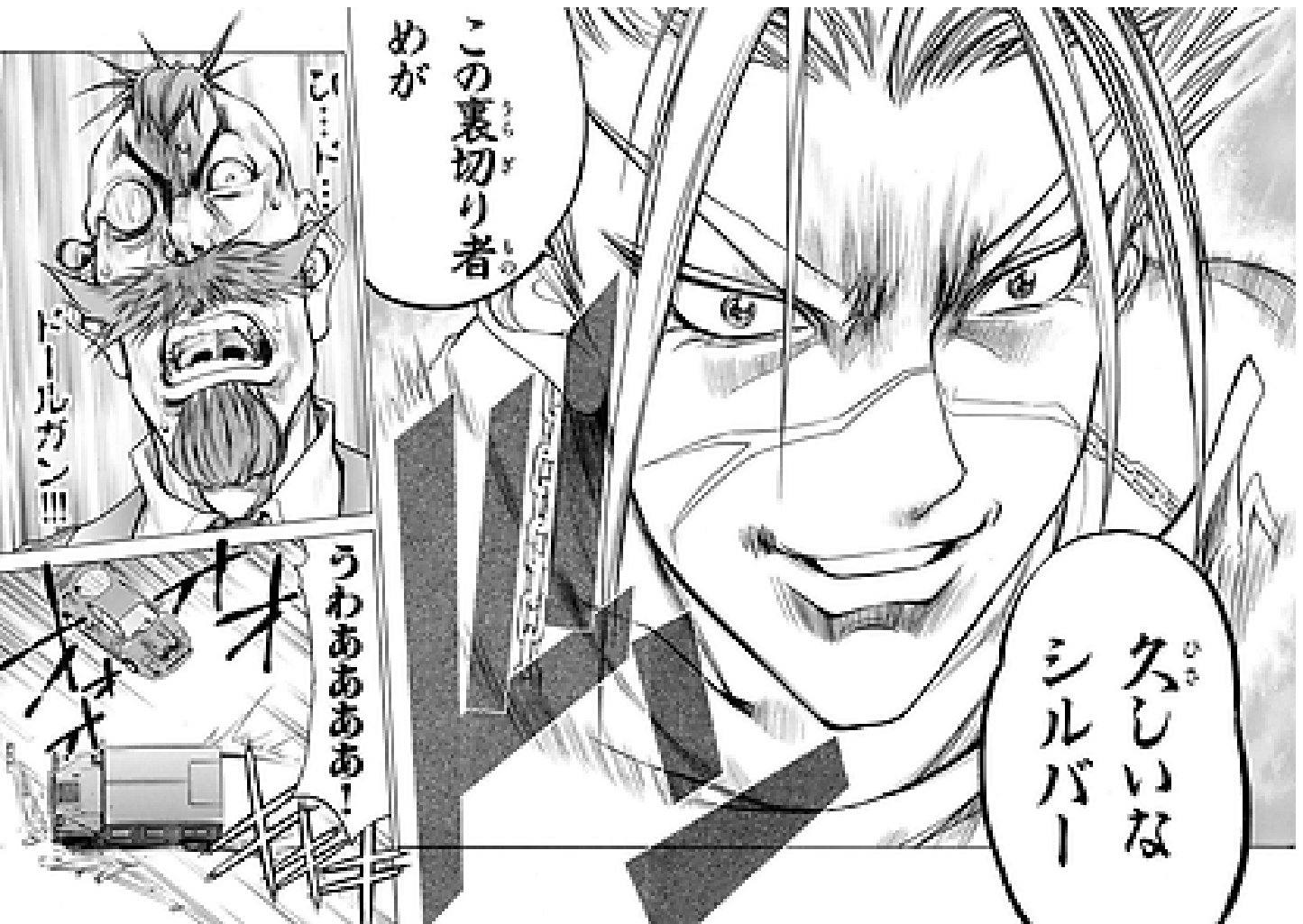}}
    \subcaption{\footnotemark[5]}
    \label{fig:dataset:double-sided:b}
  \end{minipage}
  \caption{Double-sided pages.
    In both of these images, the left and right pages are combined to describe a large image (\subref{fig:dataset:double-sided:a}).
    The character in the middle is located across both sides (\subref{fig:dataset:double-sided:b}).
    The right frame is extended to the left page.}
  \label{fig:dataset:double-sided}
\end{figure}

\begin{figure}[!t]
  \centering
  \begin{minipage}{0.32\hsize}
    \frame{\includegraphics[width=1\hsize]{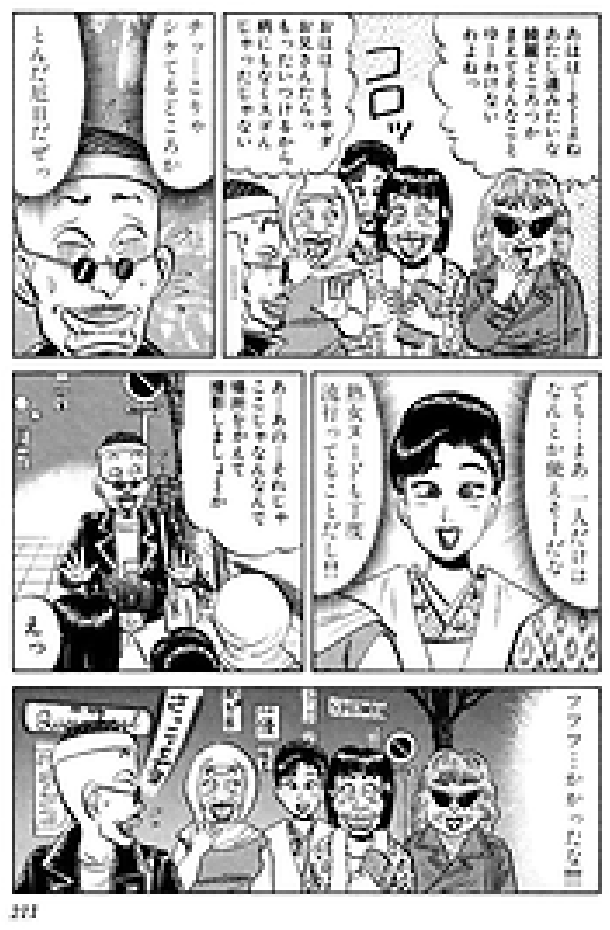}}
    \subcaption{Regular page \footnotemark[6]}
    \label{fig:dataset:irregular:regular}
  \end{minipage}
  \begin{minipage}{0.32\hsize}
    \frame{\includegraphics[width=1\hsize]{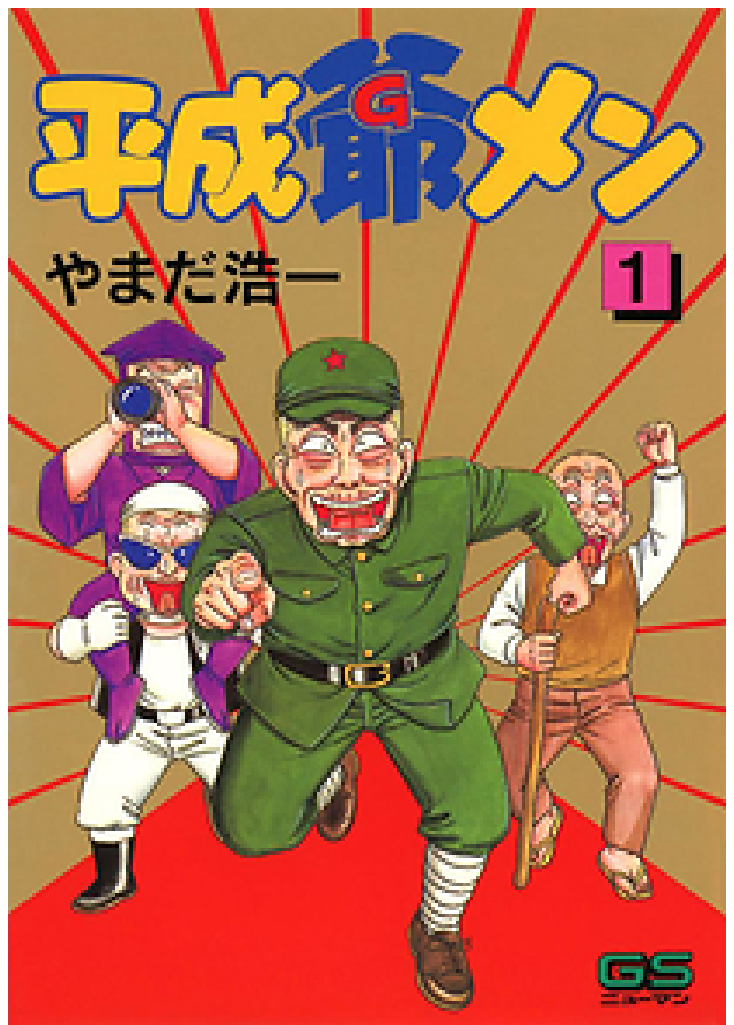}}
    \subcaption{Cover page \footnotemark[6]}
    \label{fig:dataset:irregular:cover}
  \end{minipage}
  \begin{minipage}{0.32\hsize}
    \frame{\includegraphics[width=1\hsize]{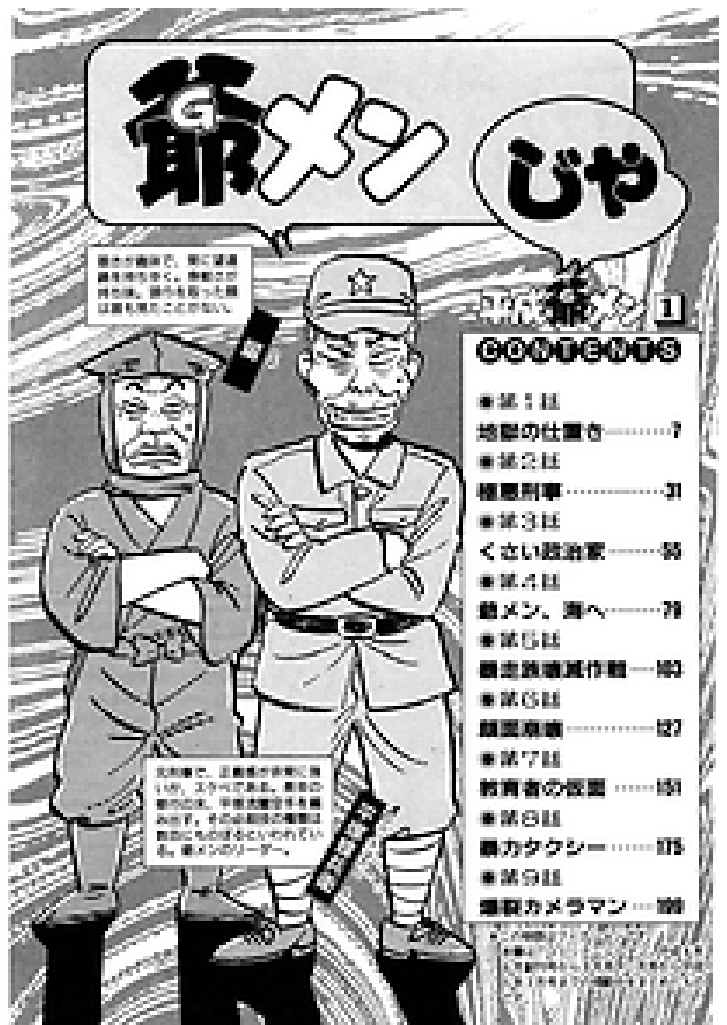}}
    \subcaption{Table of content \footnotemark[6]}
    \label{fig:dataset:irregular:toc}
  \end{minipage}
  \caption{A regular page (\subref{fig:dataset:irregular:regular}) and irregular pages (\subref{fig:dataset:irregular:cover}, \subref{fig:dataset:irregular:toc}).
    (\subref{fig:dataset:irregular:regular}) This page is separated into some frames,
    and characters and speech balloons are placed in the frames.
    Note that the letters of each text are put in vertical lines.
    This is one of the features of Japanese comics.
    (\subref{fig:dataset:irregular:cover}, \subref{fig:dataset:irregular:toc})
    There is no frame and objects are placed over whole pages.
    In this page, the letters are put in horizontal lines.}
  \label{fig:dataset:irregular}
\end{figure}

Let us describe the procedure of our manual annotation process.
Before the annotation, we conducted two preprocesses over all images.
First, we concatenated the left and right pages into one.
Sometimes, the authors combine left and right pages to describe a large image.
We call this pair of pages \textit{double-sided page} (\cref{fig:dataset:double-sided}).
In such cases, objects can be located across left and right pages.
To annotate these objects properly, we treated all pages as double-sided pages.
Note that this concatenation was conducted without considering the contents of pages.
Therefore a pair of independent pages is also concatenated.
The size of a concatenated page is typically 1654 $\times$ 1170 px.
Secondly, we skipped irregular pages, including cover pages, tables of contents, and the afterword.
The structure of these pages are different from other pages and it is difficult to treat them in the same way (\cref{fig:dataset:irregular}).
After these preprocesses, the number of pages to be annotated was 10,130 (\cref{table:dataset}).

\footnotetext[4]{``HanzaiKousyouninMinegishiEitarou'' \textcopyright Ki Takashi}
\footnotetext[5]{``DollGun'' \textcopyright Deguchi Ryusei}

The annotation was conducted in three steps.
First, we hired 72 workers.
Each worker was assigned one volume first.
If a worker performed well in terms of annotation speed,
we assigned additional volumes to the worker.
The best worker has annotated 11 volumes.
For this annotation task, we developed a cloud-based annotation tool.
As the task is complicated (selecting a region, assigning a category label, typing contents of texts, etc.),
we decided not to use Amazon Mechanical Turk, which is a famous cloud sourcing platform.
The whole annotation process took six months.
Secondly, to ensure the quality of the annotations,
we double checked all pages manually with the help of 26 workers.
The workers flagged pages with errors that were caused by mistakes and misunderstanding of annotation criteria.
As comics are complicated documents, there were many errors after the first annotation.
This double-checking step took two weeks, and 1,227 pages were marked as having errors.
Finally, we refined the annotations in the flagged pages.
This refinement was also conducted by the 26 workers who participated in the double-checking step.

\footnotetext[6]{``HeiseiJimen'' \textcopyright Yamada Koichi}

\Cref{table:dataset} shows the statistics of the Manga109-annotations.
Manga109-annotations is the largest manually annotated dataset.
In particular, unlike COMICS~\cite{IyyerComics2016},
Manga109-annotations has \textit{face} and \textit{body} annotations.
This is a clear advantage because characters would be
one of the most important objective for the detection task.

\section{Proposed detection method: SSD300-fork}
\label{sec:method}
In this section, we propose \textit{SSD300-fork}, a detection method for highly overlapped objects.
We first explain the assignment problem that is caused by highly overlapped objects.
Then we present our method, SSD300-fork to solve the assignment problem.

\subsection{Assignment problem}
We focus on the assignment problem of highly overlapped objects (\cref{fig:relatedwork:dense}),
where the problem is defined as follows.
As denoted in \cref{sec:related:natural-detection},
the state-of-the-art object detection methods employ the anchor-based approach, which is visualized in \cref{fig:method:assign:sparse}.
The dashed rectangles represent anchor boxes and the solid rectangle containing them represents an anchor set.
Each anchor box has its own size, shape, and location.
In the testing phase, given an image with several objects,
anchor-based methods predict the location and category of the nearest object for each anchor box.
To suppress the outputs that indicate the same object, non-maximum suppression is conducted during post-processing.
In the training phase, each anchor box is trained to return the location and category of the nearest object.
To achieve this, anchor-based methods assign the ground truth objects to anchor boxes.
For example, the top-right ellipse is assigned to the top-right landscape anchor box (\cref{fig:method:assign:sparse}).
This assignment is based on both spatial position and the shape of the target object.

The problem of existing methods is that they cannot assign objects to anchor boxes properly
if there are several objects of similar location, size, and aspect ratio (\cref{fig:method:assign:dense}).
Here, the blue star and the green triangle are objects of similar shape, and are placed at the same position.
Although both objects should be assigned to the bottom right square anchor box,
the assignment fails because one anchor box can handle only one object.
As the result, one of those objects is not used for training.
This becomes serious especially when the target objects are highly overlapped.
Such overlap does not usually happen for the dataset used in the literature (PASCAL VOC or MS COCO),
whereas it can often happen for the detection on other domains such as our comics dataset.
In such a case, the detection would not work well even if the number of category labels was small.

\subsection{Forked model}
\begin{figure*}[!t]
  \centering
  \begin{minipage}{0.3\hsize}
    \includegraphics[width=1\hsize]{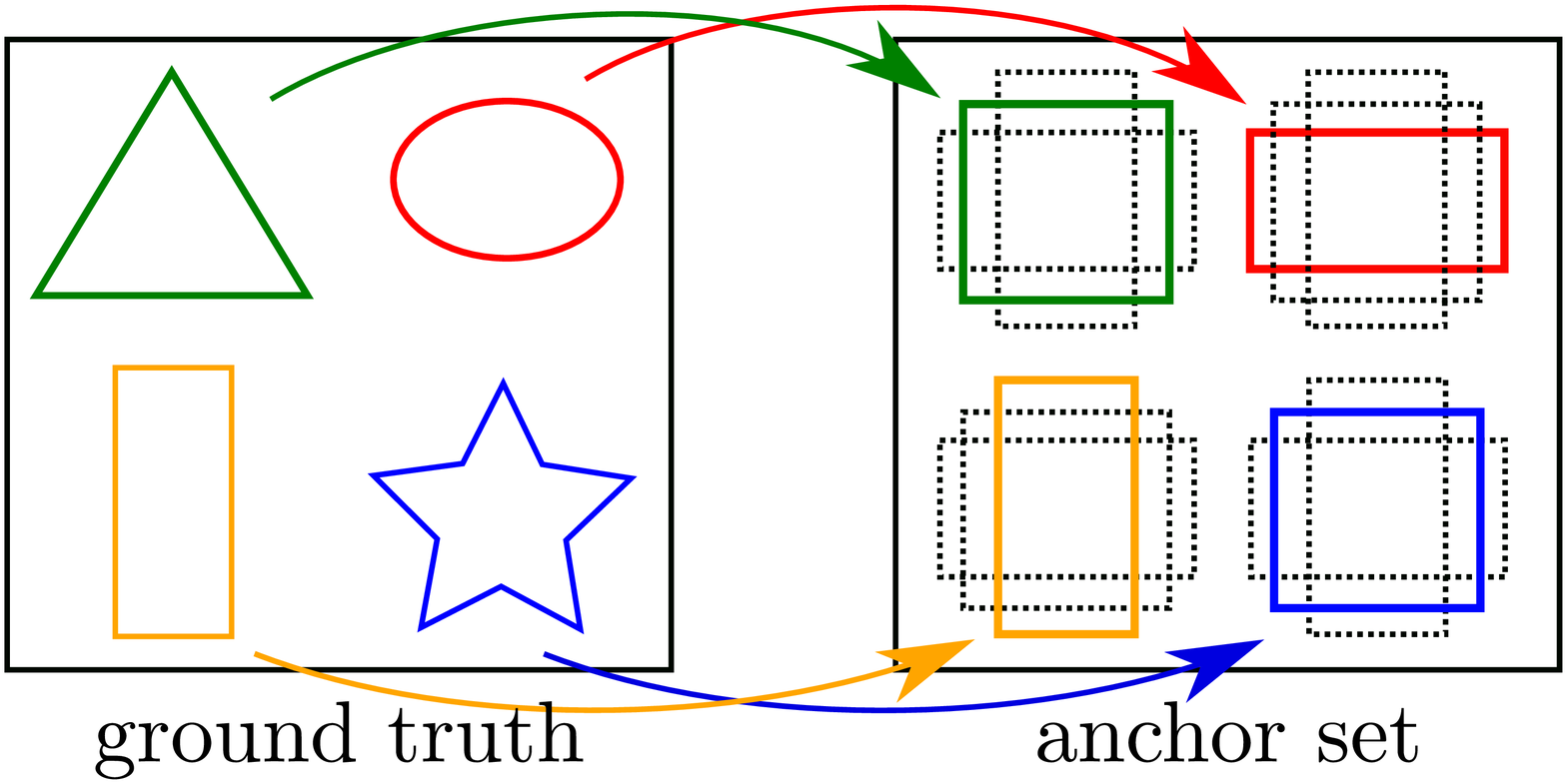}
    \subcaption{Normal images}
    \label{fig:method:assign:sparse}
  \end{minipage}
  \hfil
  \begin{minipage}{0.3\hsize}
    \includegraphics[width=1\hsize]{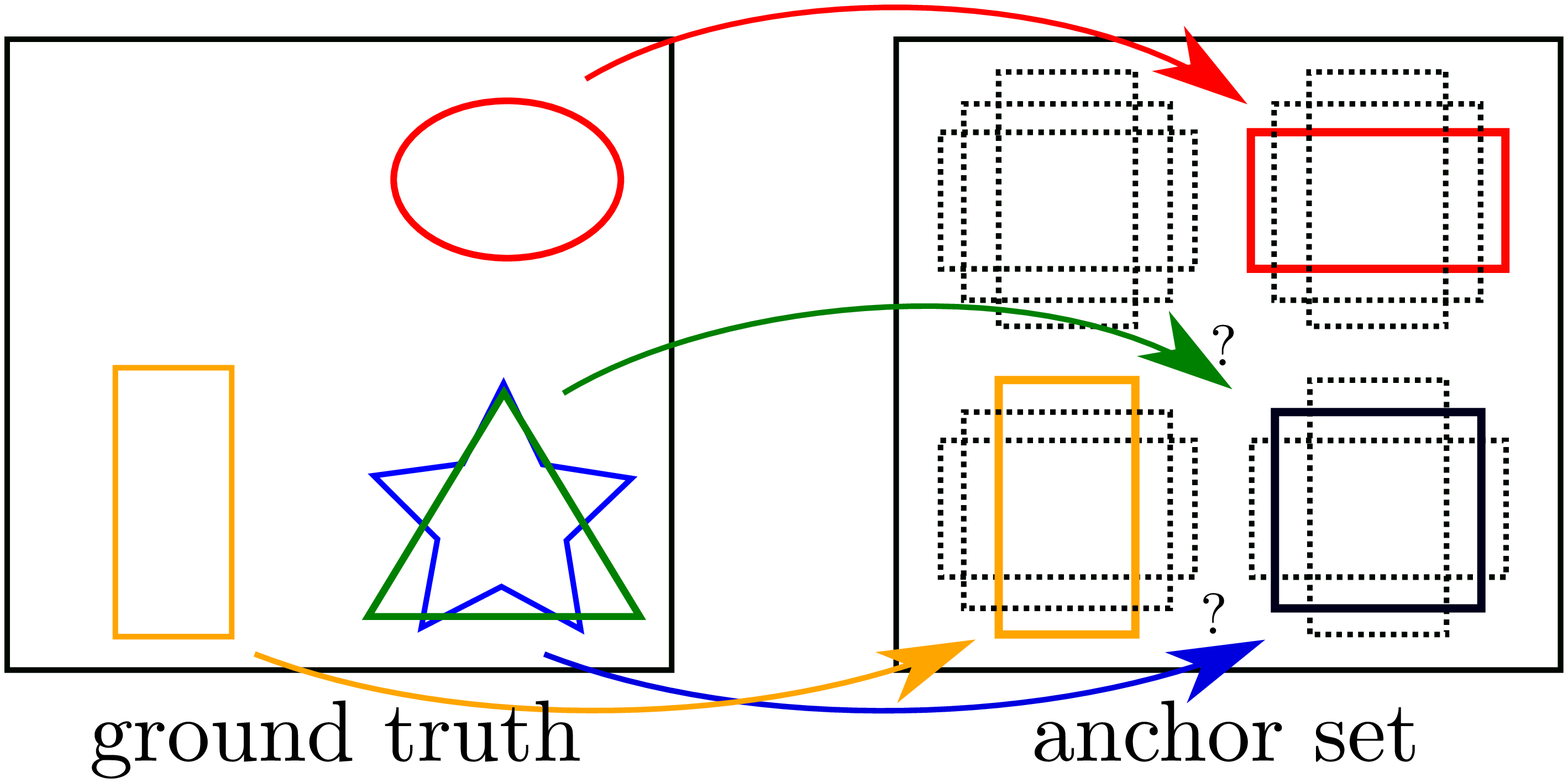}
    \subcaption{Highly overlapped objects}
    \label{fig:method:assign:dense}
  \end{minipage}
  \hfil
  \begin{minipage}{0.3\hsize}
    \includegraphics[width=1\hsize]{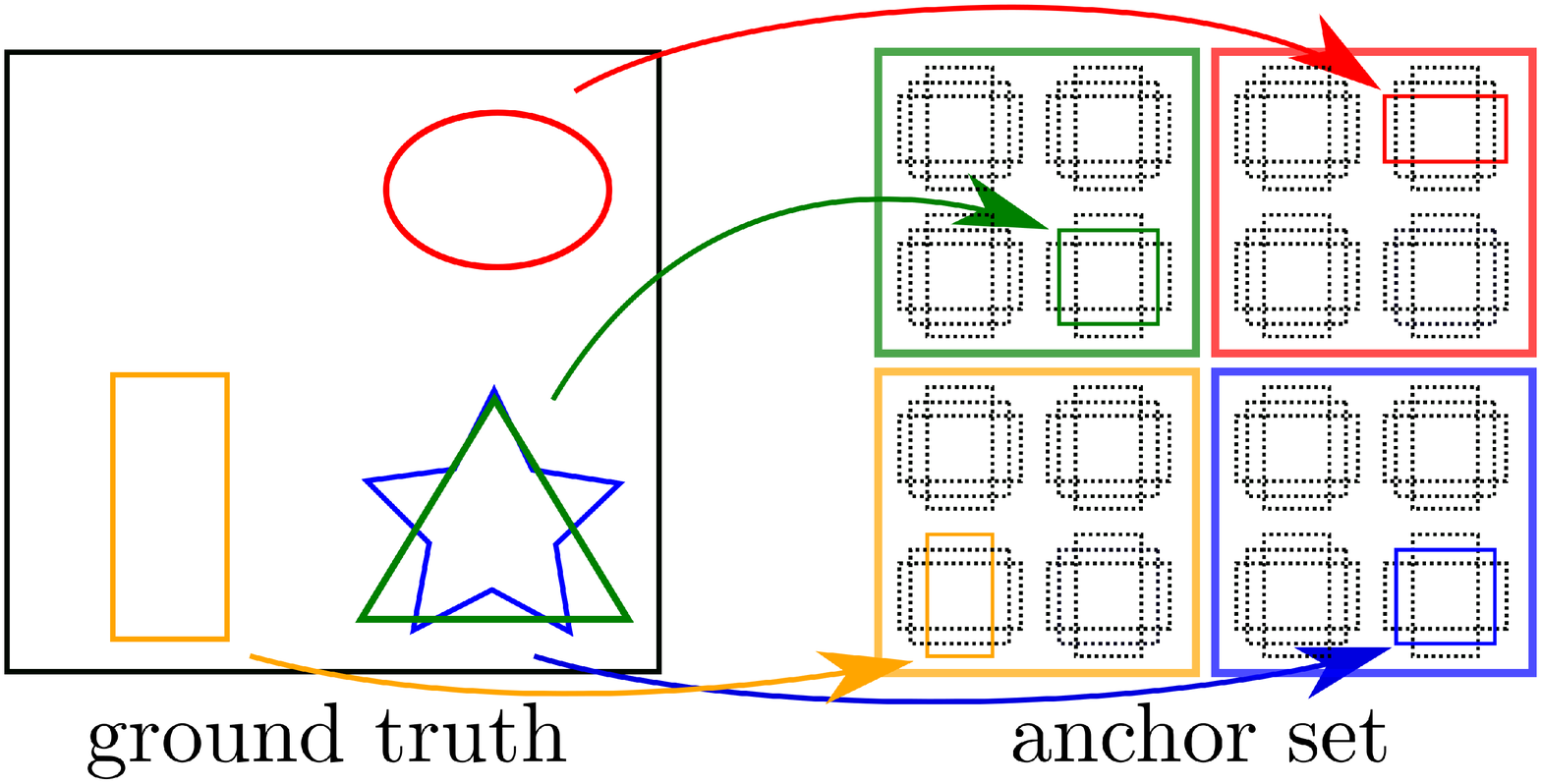}
    \subcaption{The proposed forked model}
    \label{fig:method:assign:fork}
  \end{minipage}
  \caption{``Assignment problem''.
    Objects are assigned to anchor boxes (dashed rectangles) according to their overlaps.
    (\subref{fig:method:assign:sparse}) In normal images, which do not contain highly overlapped objects,
    all objects are assigned to at least one anchor box.
    (\subref{fig:method:assign:dense}) Highly overlapped objects conflict and
    some of them are not assigned to any anchor boxes.
    In this case, a blue star and a green triangle are conflicting.
    (\subref{fig:method:assign:fork}) The proposed forked model can treat overlapped objects using replicated anchor boxes.
    The blue star and the green triangle are assigned to different anchor boxes.}
  \label{fig:method:assign}
\end{figure*}

To solve this assignment problem, we propose \textit{forked} model.
We make $C$ replicates of the anchor set, where $C$ is the number of categories
($C=4$, \{\textit{frame}, \textit{text}, \textit{face}, and \textit{body}\}).
Each replicated anchor set takes responsibility for one category.
The objects are assigned to the anchor boxes in the corresponding anchor set.
\Cref{fig:method:assign:fork} shows an example of the proposed forked model.
The blue star is assigned to the anchor box for the star category (the bottom-right anchor set),
and the green triangle is assigned to that for the triangle category  (the top-left anchor set).
By using the proposed forked model, we can solve the assignment problem,
and highly overlapped objects can be correctly detected.

\begin{figure*}[!t]
  \centering
  \begin{minipage}{0.45\hsize}
    \includegraphics[width=1\hsize]{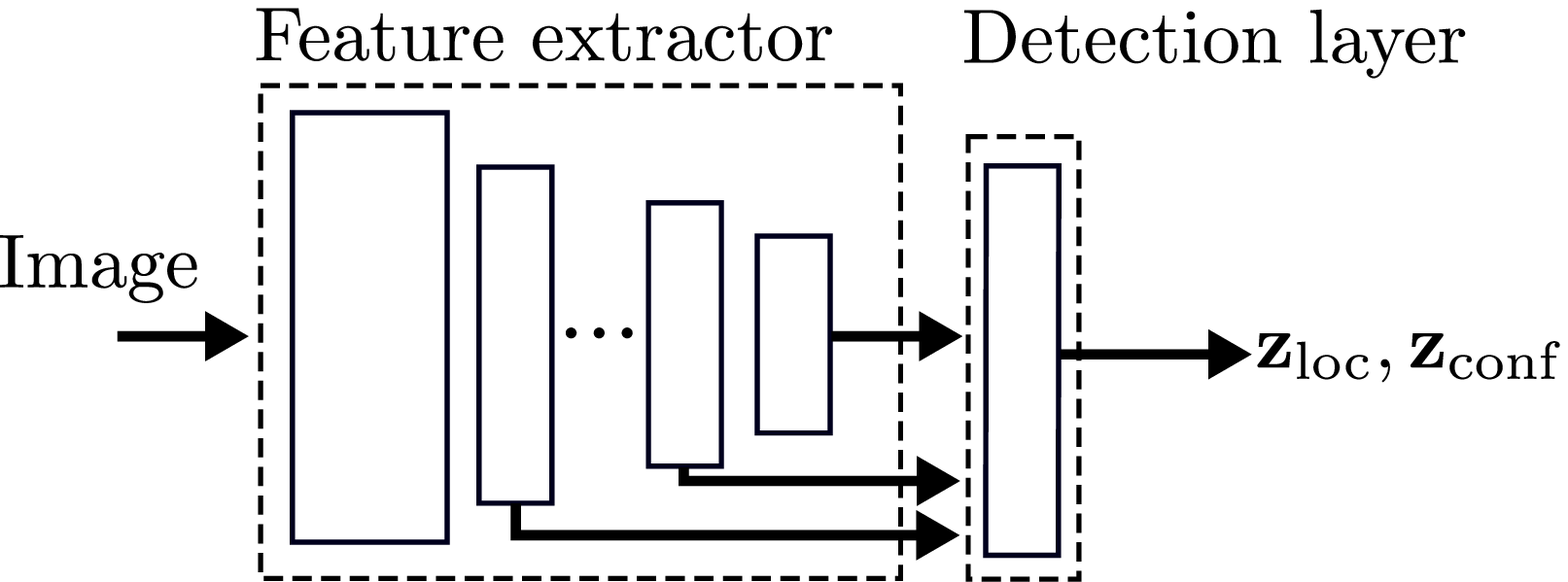}
    \subcaption{SSD300}
    \label{fig:method:architecture:orig}
  \end{minipage}
  \hfil
  \begin{minipage}{0.45\hsize}
    \centering
    \includegraphics[width=1\hsize]{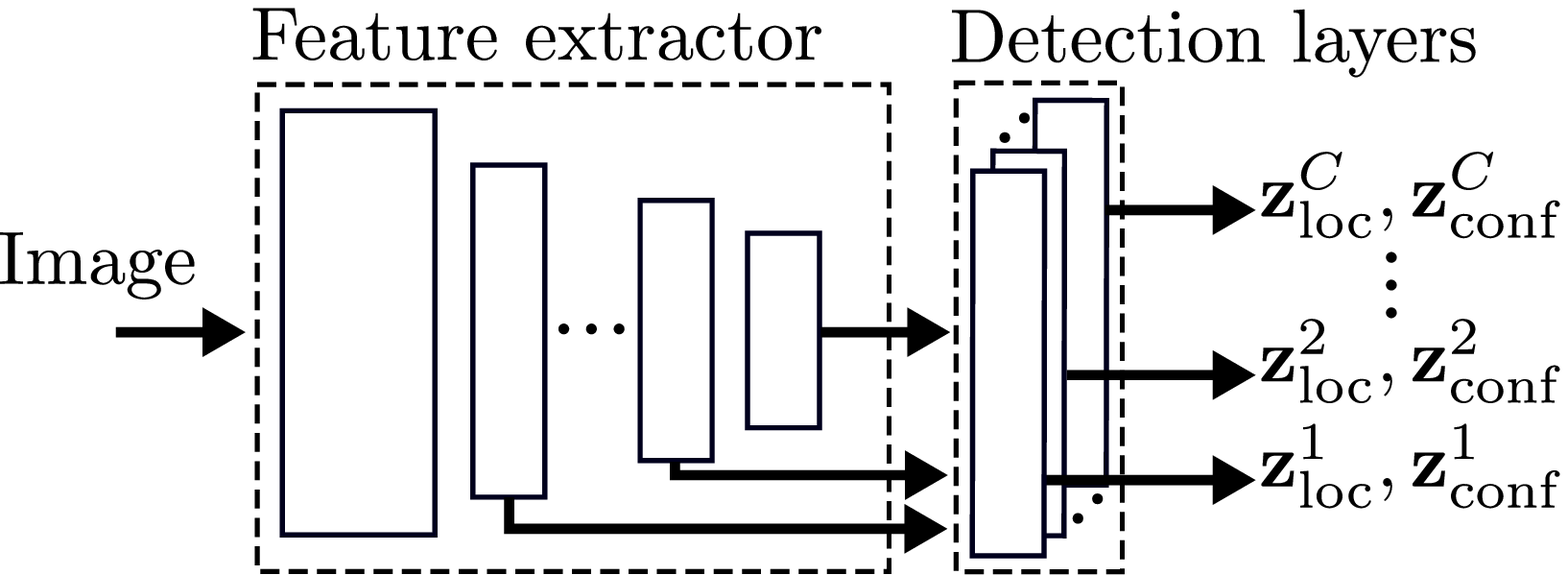}
    \subcaption{SSD300-fork}
    \label{fig:method:architecture:fork}
  \end{minipage}
  \caption{Network architectures.
    (\subref{fig:method:architecture:orig}) SSD300 comprises two parts: a feature extractor and a detection layer.
    The detection layer takes multi-scale feature maps and returns two prediction tensors:
    ${\bf z}_{\rm loc}\in\mathbb R^{K\times4}$ and ${\bf z}_{\rm conf}\in\mathbb R^{K\times(C+1)}$,
    where $K$ is the number of anchor boxes ($K=8732$) and $C$ is the number of categories ($C=4$).
    (\subref{fig:method:architecture:fork}) SSD300-fork has a shared feature extractor and $C$ replications of the detection layer.
    The $c$-th detection layer returns $\mathbf{z}_\mathrm{loc}^c\in\mathbb R^{K\times4}$ and $\mathbf{z}_\mathrm{conf}^c\in\mathbb R^K$.}
  \label{fig:method:architecture}
\end{figure*}

To build the forked model, we chose SSD300~\cite{liu2016ssd} as the base network of our method
because it archived the best performance among existing CNN-based detection methods (\cref{table:experiment:result}).
SSD300 is made of a multi-scale feature extractor and a detection layer (\cref{fig:method:architecture:orig}).
The multi-scale feature extractor is a convolutional neural network
that takes an image and returns a set of features maps.
The detection layer takes a set of feature maps computed by the feature extractor
and returns two tensors: ${\bf z}_{\rm loc}$ and ${\bf z}_{\rm conf}$.
Note that $\mathbf{z}_\mathrm{loc}\in\mathbb R^{K\times4}$ indicates the locations of objects,
where $K$ is the number of anchor boxes and
each anchor box has four values ($\Delta x_\mathrm{center},\Delta y_\mathrm{center},\Delta width,\Delta height$).
These four values represent the offset between the predicted bounding box and the anchor box.
In the same manner, $\mathbf{z}_\mathrm{conf}\in\mathbb R^{K\times(C+1)}$ indicates the categories of objects and
each category label is represented by $(C+1)$ values.
The $c$-th value of the $(C+1)$ values indicates
the softmax probability that the region of the anchor box belongs to the $c$-th category.
The $(C+1)$-th value is a special value that indicates the probability of the background (not belonging to any categories).
The number of the anchor boxes, $K=8732$, is defined by $K=\sum_{i=1}^Fk_ig_i^2$,
where $F$ is the number of feature maps ($F=6$),
$k_i$ is the number of shapes of anchor boxes for the $i$-th feature map ($k_1=4,k_2=6,k_3=6,k_4=6,k_5=4,k_6=4$),
and $g_i$ is the size of the $i$-th feature map ($g_1=38,g_2=19,g_3=10,g_4=5,g_5=3,g_6=1$).

Our method, SSD300-fork, has a shared feature extractor and $C$ replicated detection layers (\cref{fig:method:architecture:fork}).
The architecture of the feature extractor is the same as that of SSD300,
and it computes a set of feature maps that is shared among all detection layers.
The $c$-th detection layer corresponds to the $c$-th category.
It returns $\mathbf{z}_\mathrm{loc}^c\in\mathbb R^{K\times4}$ and $\mathbf{z}_\mathrm{conf}^c\in\mathbb R^K$.
Each value of $\mathbf{z}_\mathrm{conf}^c$ indicates
the sigmoid probability that the region of the anchor box belongs to the $c$-th category.
Note that we can replicate the anchor set more simply by replicating the entire network.
However, this approach requires four times the number of parameters and training/testing time.
By forking only the final layer, we reduce the number of parameters and the computation cost.
The number of parameters is 25.6 M, while that of SSD300 is 24.1 M and that of na{\"i}ve approach is 95.0 M.
The training/testing time of SSD300-fork is kept almost the same as that of SSD300 (about 10 FPS).

For SSD300-fork, we introduce a weighted category-wise loss to balance the difficulties of categories.
This loss function is defined as follows:
\begin{align}
  L\left({\bf z}_{\rm loc}^1,{\bf z}_{\rm conf}^1,\dots,{\bf z}_{\rm loc}^C,{\bf z}_{\rm conf}^C\right)
  =\sum_{c=1}^Cw_c\frac{L_{\rm loc}\left({\bf z}_{\rm loc}^c\right)+L_{\rm conf}\left({\bf z}_{\rm conf}^c\right)}{N_+^c},
\end{align}
where $L_{\rm loc}$ is the localization loss that is also used in the original SSD300.
$L_{\rm conf}$ is the classification loss.
This loss is the same as the original one, except for using sigmoid cross entropy instead of softmax cross entropy.
The hard-negative mining process, which Liu et al.~\cite{liu2016ssd} introduced for stabilization,
is also conducted.
$N_+^c$ is the number of anchor boxes that are assigned to groundtruth objects.
$w_c$ is the weight of the $c$-th categories. Experimentally, we set $(w_1,w_2,w_3,w_4) = (0.2, 0.2, 0.4, 0.2)$.
$w_1$, $w_2$, $w_3$, and $w_4$ are the weights of \textit{frame}, \textit{text}, \textit{face}, and \textit{body} respectively.

\section{Experiment}
We evaluated our SSD300-fork using Manga109-annotations and eBDtheque~\cite{eBDtheque2013}.

\subsection{Comparison with CNN-based methods}
We compared SSD300-fork with other CNN-based object detection methods using the Manga109-annotations dataset.
Thanks to the large scale annotated dataset,
we easily trained, evaluated, and compared machine learning-based methods.
For all existing methods, we used the configurations defined for PASCAL VOC.
We used the original implementation for YOLOv2 and the implementations provided by ChainerCV~\cite{ChainerCV2017} for Faster~R-CNN and SSD300.

For SSD300-fork, we followed the configurations of SSD300.
The weights of feature extractor were initialized with those of VGG-16~\cite{simonyan2015very} trained on ImageNet~\cite{deng2009imagenet}.
For the detection layers, we made use of the uniform distribution proposed by LeCun~et al.~\cite{lecun2012efficient}.
We used all preprocessing methods in our SSD300-fork that were used in SSD300, such as color augmentation, random expansion, random cropping, resizing, flipping, and mean subtraction.
Mean values of ImageNet are used for the random expansion and the mean subtraction in SSD300-fork.
The learning rate was set to $10^{-3}$ initially,
and it was changed to $10^{-4}$ and $10^{-5}$ at the 80,000th and 100,000th iterations respectively.
We stopped the training at 120,000 iterations (=3,791 epochs).
This learning rate strategy is the same as SSD300 used for PASCAL VOC~\cite{Everingham15}.
In addition, the learning rate for the detection layers were multiplied by four
because the number of parameters of these layers were increased by replication.
We implemented SSD300-fork using Chainer~\cite{chainer_learningsys2015} and ChainerCV.

\Cref{table:experiment:result} shows average precision (AP) scores and their mean over the four categories, i.e., mean average precision (mAP).
We followed PASCAL VOC metrics and used ${\rm IoU}\geq0.5$ as the threshold,
which is standard criteria in object detection for naturalistic images.
SSD300-fork outperformed other methods.
In particular, the AP for \textit{face} was gained by 9~\% against the original SSD300.

\footnotetext[7]{``YukiNoFuruMachi'' \textcopyright Yamada Uduki}

\begin{figure}[!t]
  \centering
  \begin{minipage}{0.4\hsize}
    \frame{\includegraphics[width=1\hsize]{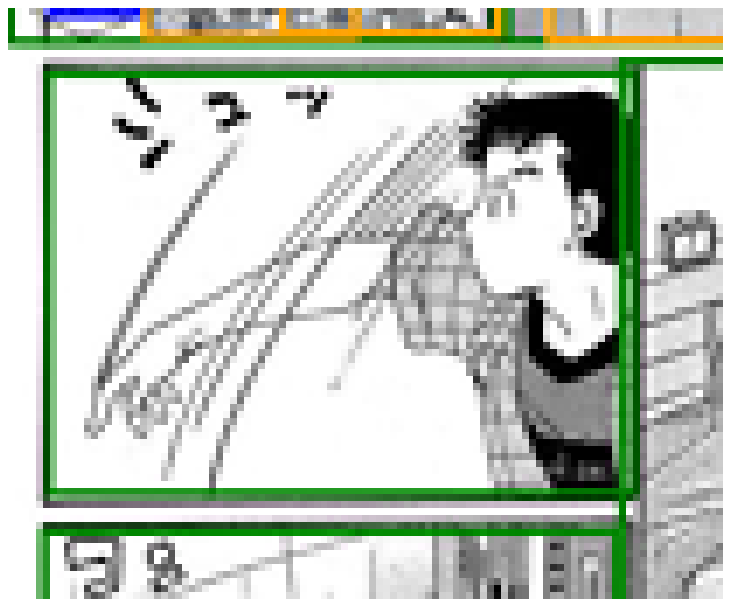}}
    \subcaption{SSD300\footnotemark[1]}
    \label{fig:experiment:compare:orig:a}
  \end{minipage}
  \hfil
  \begin{minipage}{0.4\hsize}
    \frame{\includegraphics[width=1\hsize]{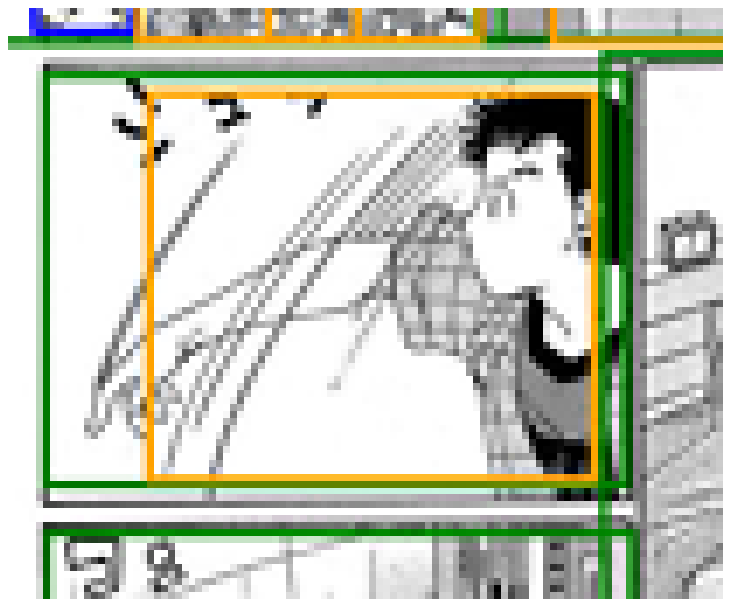}}
    \subcaption{SSD300-fork\footnotemark[1]}
    \label{fig:experiment:compare:fork:a}
  \end{minipage} \\
  \begin{minipage}{0.4\hsize}
    \frame{\includegraphics[width=1\hsize]{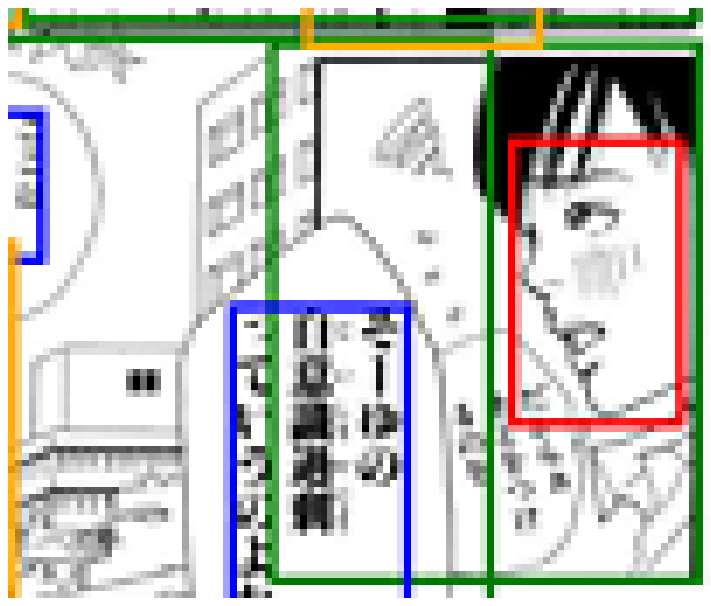}}
    \subcaption{SSD300\footnotemark[7]}
    \label{fig:experiment:compare:orig:b}
  \end{minipage}
  \hfil
  \begin{minipage}{0.4\hsize}
    \frame{\includegraphics[width=1\hsize]{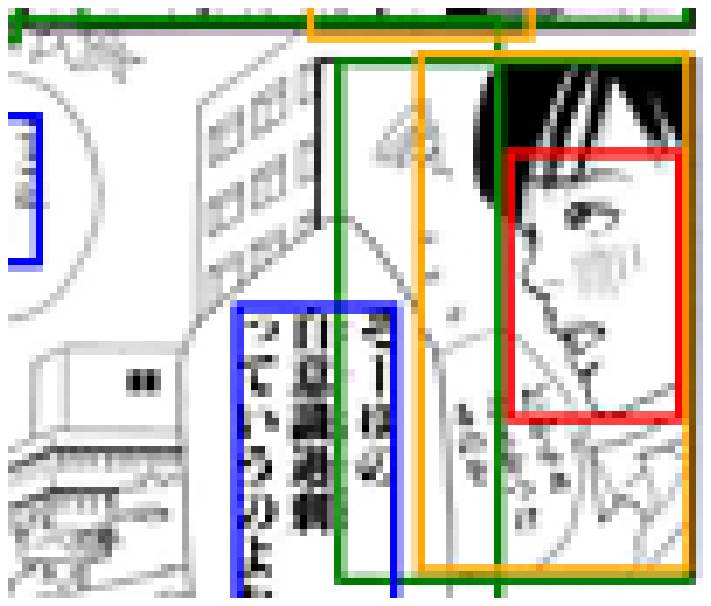}}
    \subcaption{SSD300-fork\footnotemark[7]}
    \label{fig:experiment:compare:fork:b}
  \end{minipage} \\
  \caption{The comparison of
    SSD300 (\subref{fig:experiment:compare:orig:a}, \subref{fig:experiment:compare:orig:b}) and
    SSD300-fork (\subref{fig:experiment:compare:fork:a}, \subref{fig:experiment:compare:fork:b}).
    In these pages, \textit{frame} (green rectangle) and \textit{body} (orange rectangle) are highly overlapped.
    SSD300-fork detected both \textit{frame} and \textit{body}, while SSD300 failed to detect \textit{body}.}
  \label{fig:experiment:compare}
\end{figure}

\Cref{fig:experiment:compare} shows examples where SSD300-fork correctly
detected highly overlapped objects.
In these examples, \textit{frame} and \textit{body} are highly overlapped.
SSD300-fork detected both categories properly,
whereas SSD300 failed to detect \textit{body}.
\begin{table}[!t]
  \caption{The comparison with methods designed for naturalistic images using Manga109-annotations.}
  \label{table:experiment:result}
  \centering
  \begin{tabular}{@{}lrrrrr@{}}
    \toprule
    &&\multicolumn{4}{c}{AP for each category} \\ \cmidrule(l){3-6}
    Method&mAP&frame&text&face&body \\ \midrule
    Faster~R-CNN~\cite{ren2015faster}&49.9&96.1&23.8&15.7&63.9 \\
    SSD300~\cite{liu2016ssd}&81.3&\bf{97.1}&82.0&67.1&79.1 \\
    YOLOv2~\cite{redmon2016yolo9000}&59.7&90.2&64.6&37.1&46.9 \\ \midrule
    SSD300-fork&\bf{84.2}&96.9&\bf{84.1}&\bf{76.2}&\bf{79.6} \\
    \bottomrule
  \end{tabular}
\end{table}

\subsection{Analysis for each volume}
\Cref{table:experiment:result-volume} shows the score of SSD300-fork for each volume.
If the author of a volume has other works in the train split, the volume is marked with $\ast$.
Because of the same author, the similar drawing style of the marked volume might be in the train split.
\Cref{table:experiment:result-volume} shows the score is not affected by whether the style of a given volume was trained or not.
\begin{table*}[!t]
  \caption{The details of 10 test volumes of Manga109-annotations and the AP score of SSD300-fork for each volume.}
  \label{table:experiment:result-volume}
  \centering
  \begin{threeparttable}
    \begin{tabular}{@{}lrlrrrrr@{}}
      \toprule
      &&&&\multicolumn{4}{c}{AP for each category} \\ \cmidrule(l){5-8}
      Volume&\#page&Genre&mAP&frame&text&face&body \\ \midrule
      UltraEleven&108&sports&84.5&95.2&88.4&79.6&75.0 \\
      UnbalanceTokyo&82&science fiction&79.8&98.6&83.3&62.8&74.4 \\
      WarewareHaOniDearu&91&romantic comedy&83.2&94.0&84.9&71.1&82.8 \\
      YamatoNoHane&109&sports&92.0&98.6&85.7&92.8&90.9 \\
      YasasiiAkuma&89&fantasy&89.3&97.8&93.5&80.6&85.2 \\
      YouchienBoueigumi&26&four-frames&83.4&100.0&73.6&77.6&82.5 \\
      YoumaKourin $\ast$&101&fantasy&83.6&99.2&91.8&74.9&68.4 \\
      YukiNoFuruMachi $\ast$&93&love romance&77.1&95.4&71.2&63.5&78.4 \\
      YumeNoKayoiji $\ast$&96&fantasy&84.4&94.2&77.9&81.4&84.2 \\
      YumeiroCooking&85&love romance&89.0&97.7&87.0&82.4&89.1 \\ \midrule
      Total&880&\dash&84.2&96.9&84.4&76.2&79.6 \\
      \bottomrule
    \end{tabular}
    \begin{tablenotes}
      \item $\ast$ The train split contains works by the same author.
    \end{tablenotes}
  \end{threeparttable}
\end{table*}

\begin{figure}[!t]
  \centering
  \frame{\includegraphics[width=0.7\hsize]{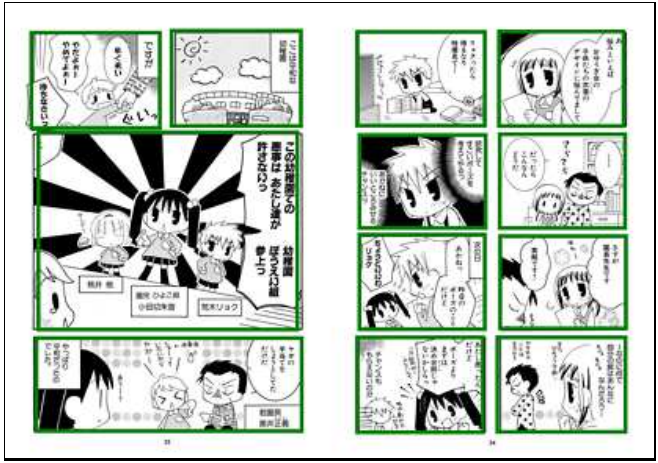}}
  \caption{\textit{frame} detection in four-frames comics \protect\footnotemark[8].
    All \textit{frame}s are detected properly.}
  \label{fig:experiment:4frame}
\end{figure}

For all volumes, the AP of \textit{frame} is higher than that of other categories.
In particular, the AP of ``YouchienBoueigumi'' is 100~\%.
The genre of this comic is four-frames comics, in which four frames with the same size are put in a vertical line (\cref{fig:experiment:4frame}).
This simple layout made \textit{frame} detection easier.
\footnotetext[8]{``YouchienBoueigumi''\textcopyright Tenya}

\begin{figure}[!t]
  \centering
  \begin{minipage}{0.35\hsize}
    \frame{\includegraphics[width=1\hsize]{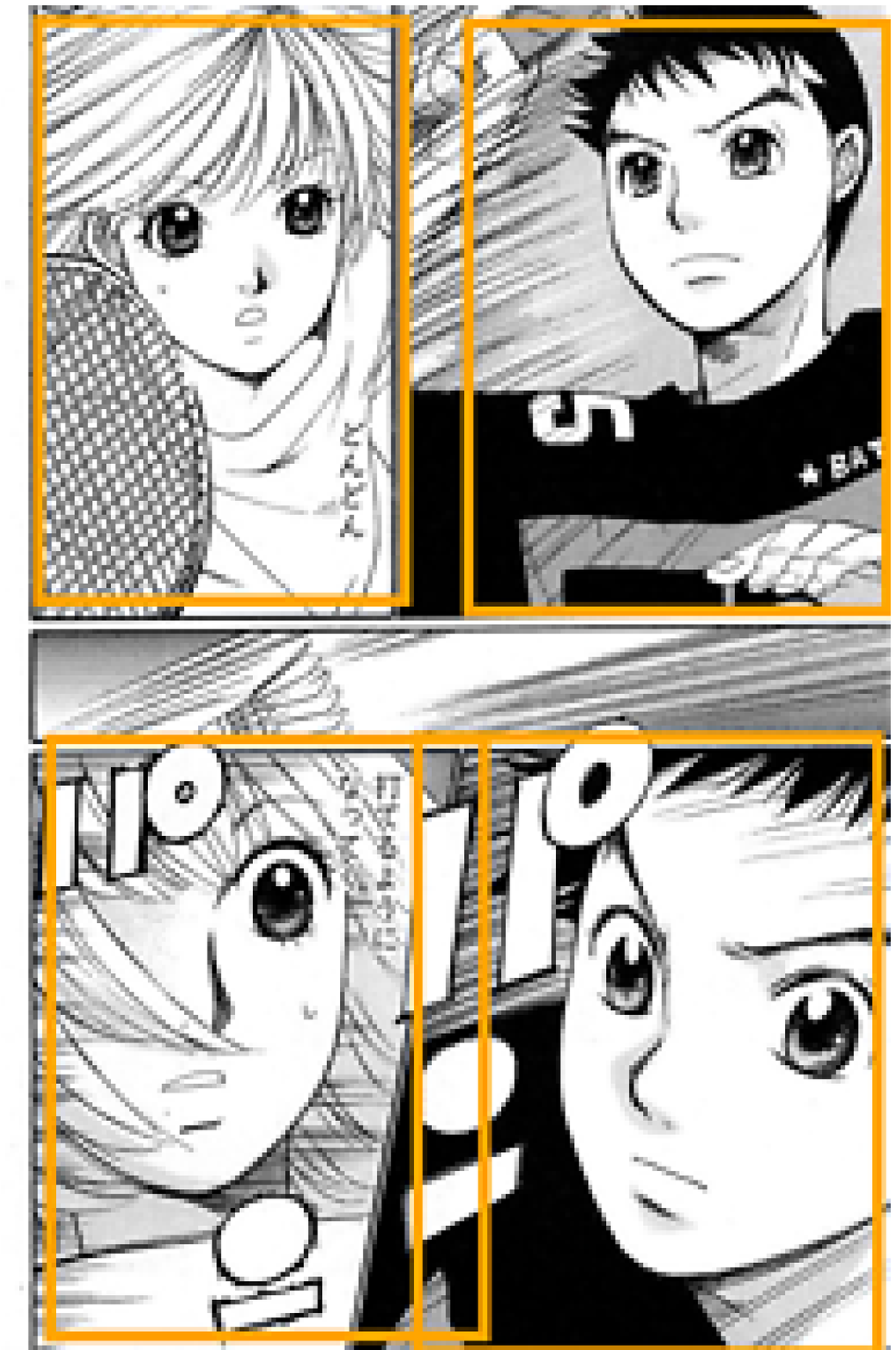}}
    \subcaption{Easy page \footnotemark[1]}
    \label{fig:experiment:body:good}
  \end{minipage}
  \hfil
  \begin{minipage}{0.35\hsize}
    \frame{\includegraphics[width=1\hsize]{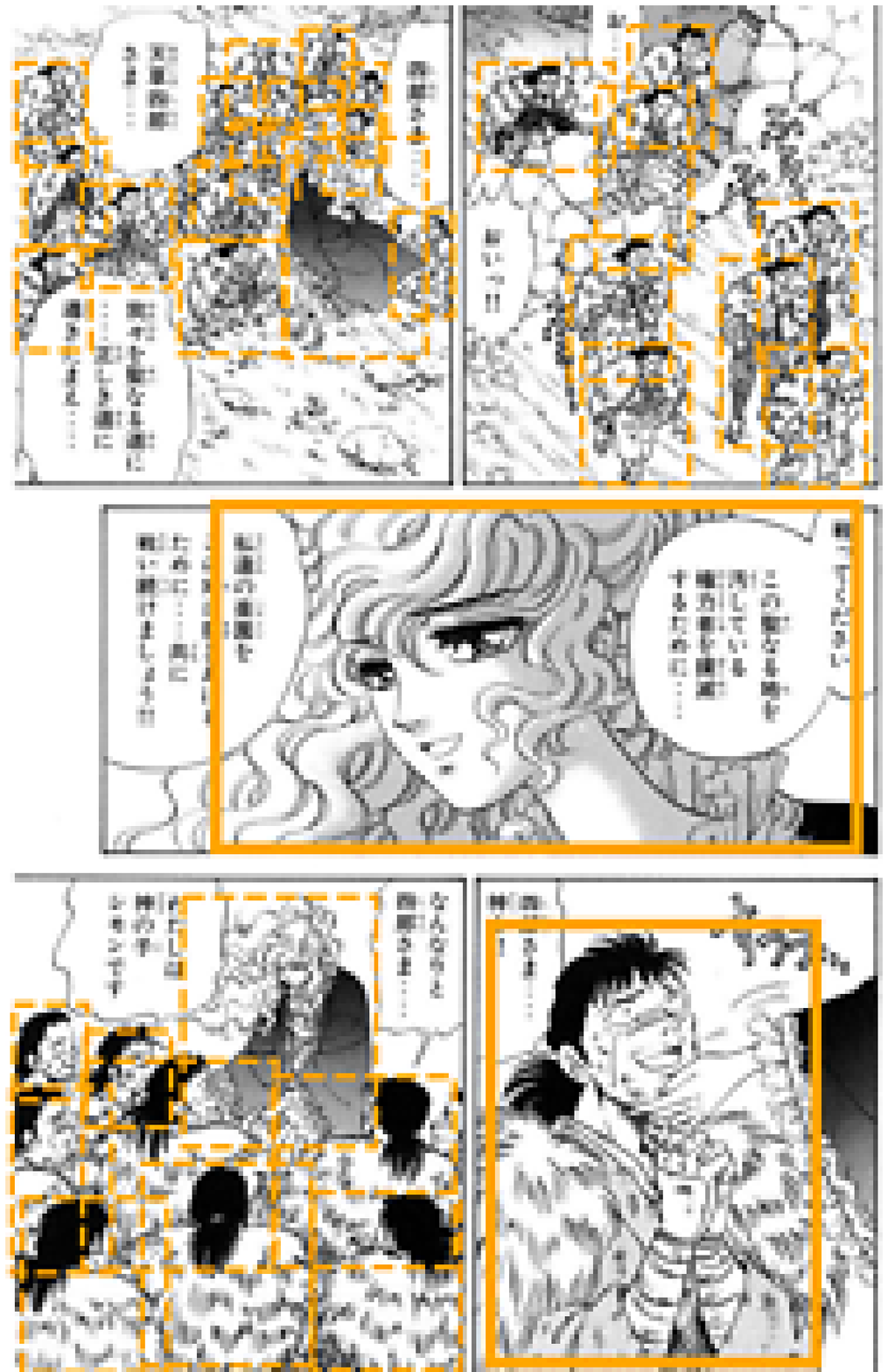}}
    \subcaption{Difficult page \footnotemark[9]}
    \label{fig:experiment:body:bad}
  \end{minipage}
  \caption{\textit{body} detection.
    We show the undetected groundtruth bodies with dashed rectangles.}
  \label{fig:experiment:body}
\end{figure}

``YamatoNoHane'' has the highest AP of \textit{body}.
This volume contains many frames with zoomed bodies to describe the actions (\cref{fig:experiment:body:good}).
On the other hand, ``YoumaKourin'' has the lowest AP.
This volume contains many frames with many tiny characters.
For example, the top and left-bottom frames of \cref{fig:experiment:body:bad} contain a lot of characters but none of them were detected.
\footnotetext[9]{``YoumaKourin'' \textcopyright Shimazaki Yuzuru, Taka Tsukasa}

\begin{figure}[!t]
  \centering
  \frame{\includegraphics[width=0.7\hsize]{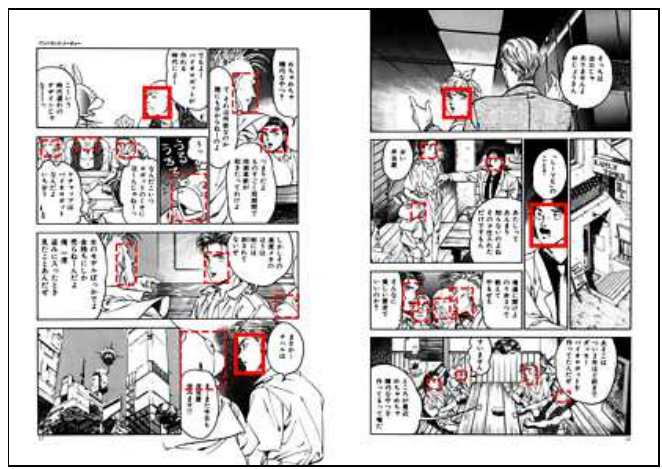}}
  \caption{\textit{face} detection \protect\footnotemark[10].
    Twenty-five faces are annotated in this page (a double-sided page).
    Only four of them were detected.
    We show the undetected groundtruth faces with dashed rectangles.}
  \label{fig:experiment:face}
\end{figure}

The most difficult volume for \textit{face} detection is ``UnbalanceTokyo''.
The faces of this volume are smaller than those of usual volumes
because the author uses a realistic style (\cref{fig:experiment:face}).
In this page, 25 faces are annotated, but only four of them were detected.

\subsection{Comparison using eBDtheque}
\begin{figure}[!t]
  \centering
  \begin{tabular}{cc}
    \frame{\includegraphics[width=0.25\hsize]{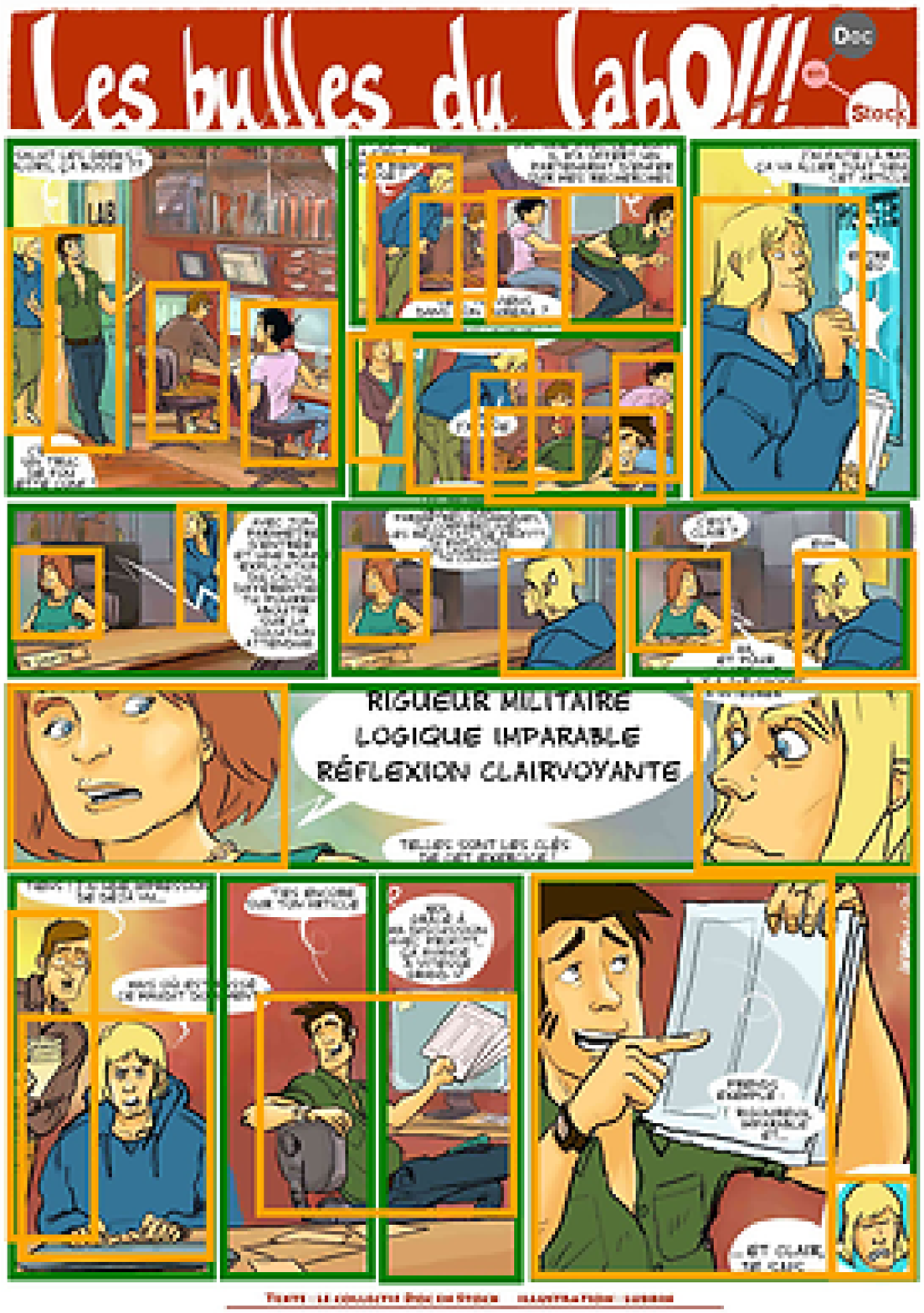}} &
    \frame{\includegraphics[width=0.25\hsize]{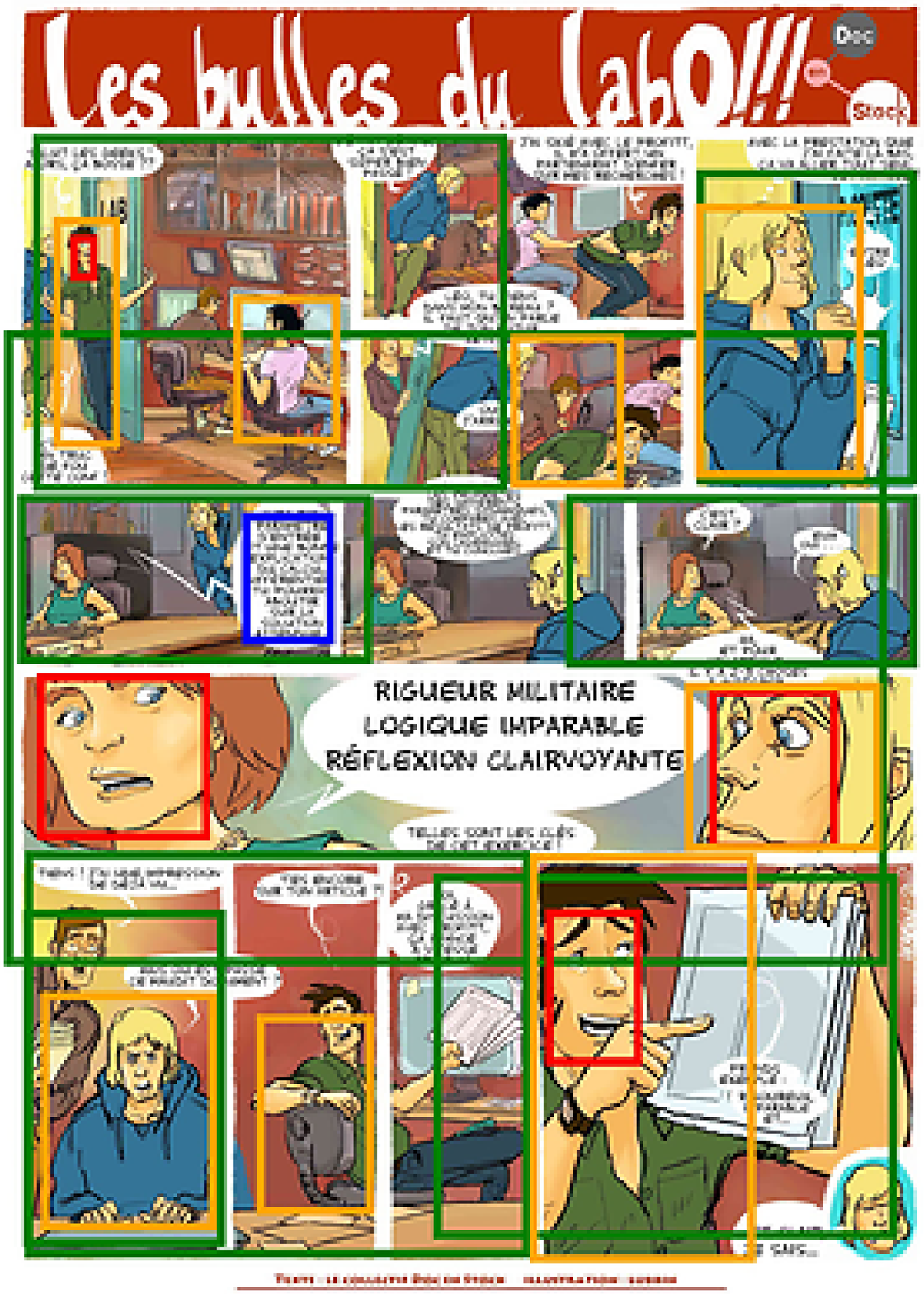}} \\
    \\
    \frame{\includegraphics[width=0.25\hsize]{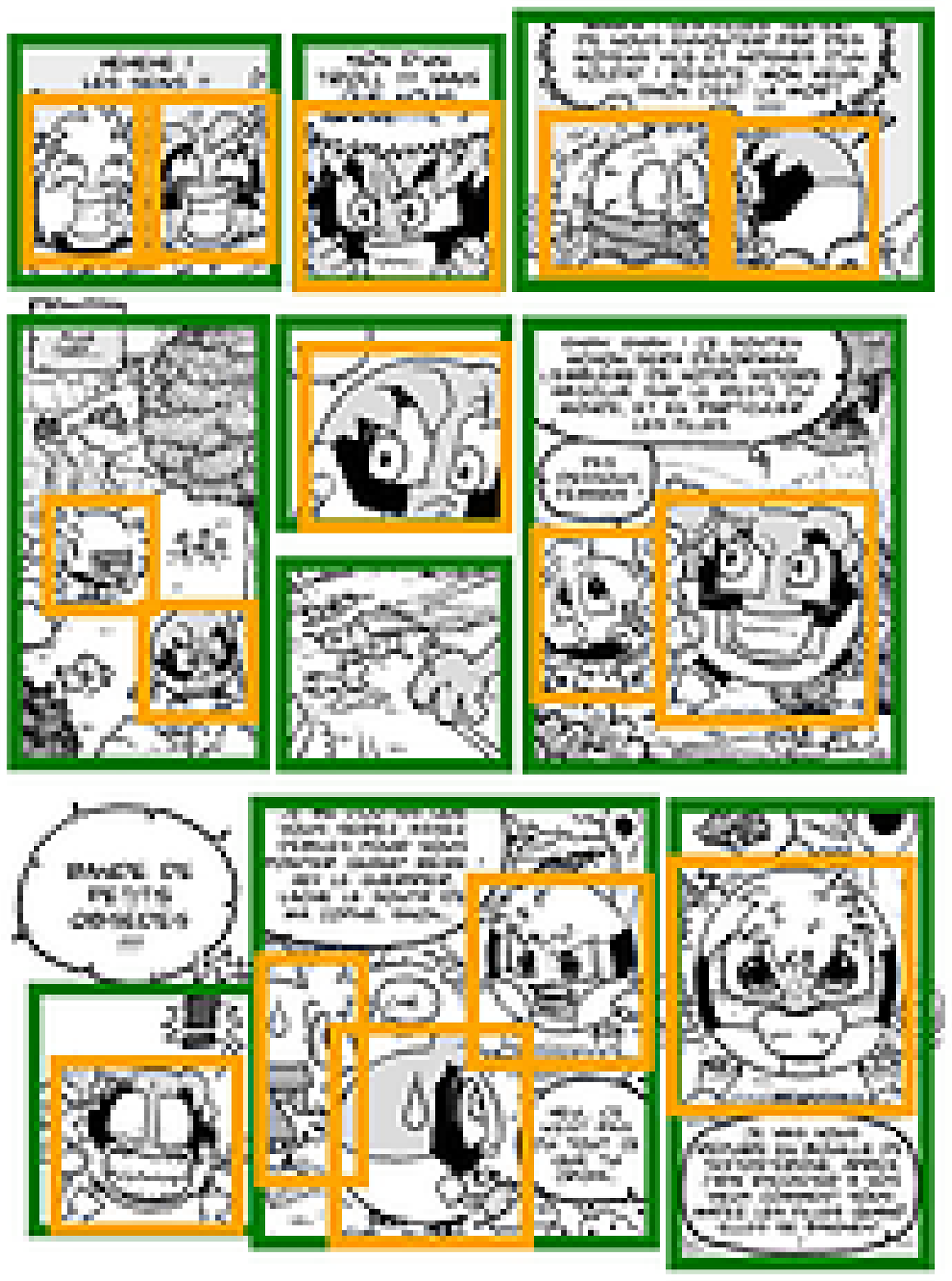}} &
    \frame{\includegraphics[width=0.25\hsize]{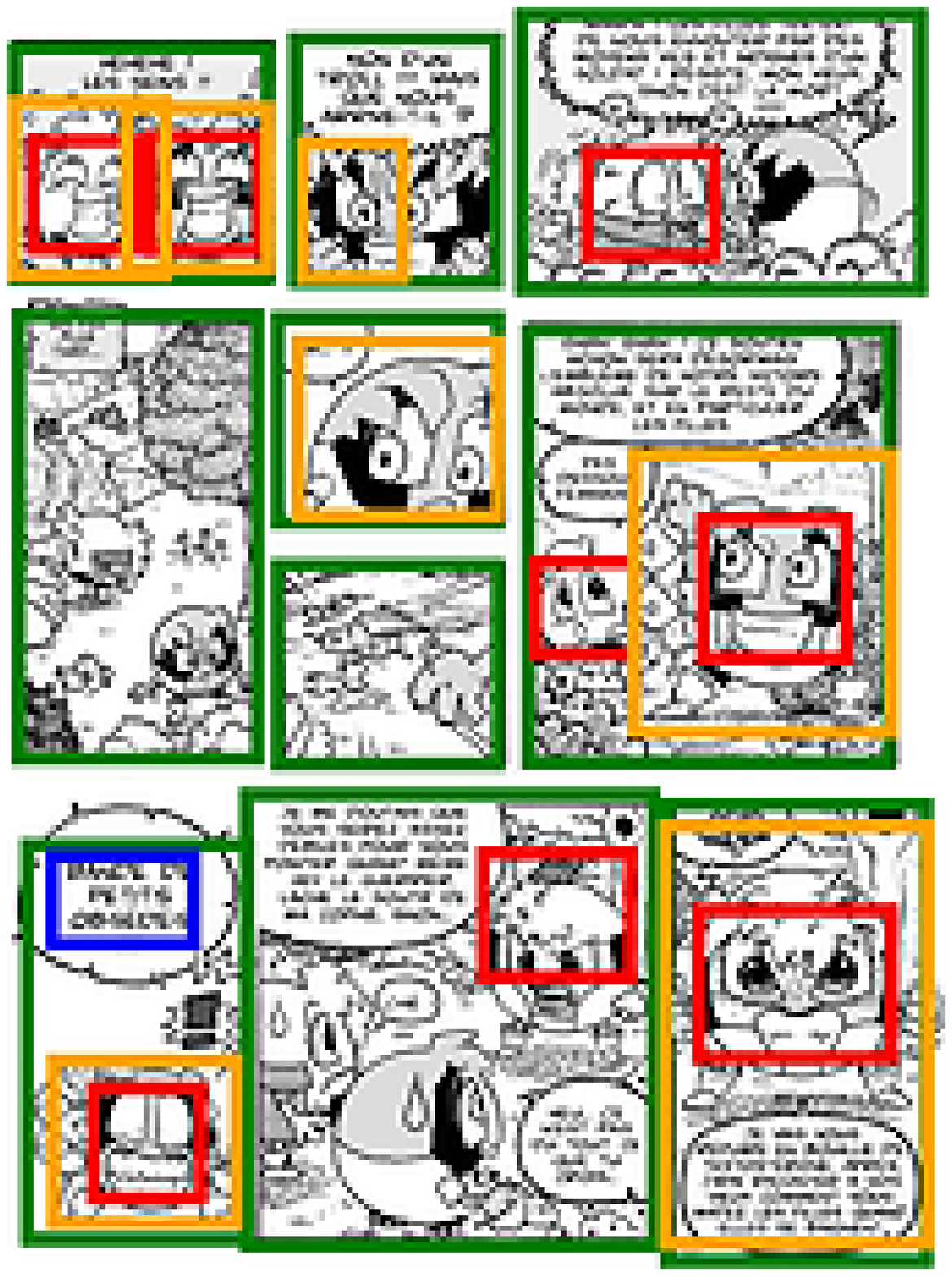}} \\
    \\
    \frame{\includegraphics[width=0.25\hsize]{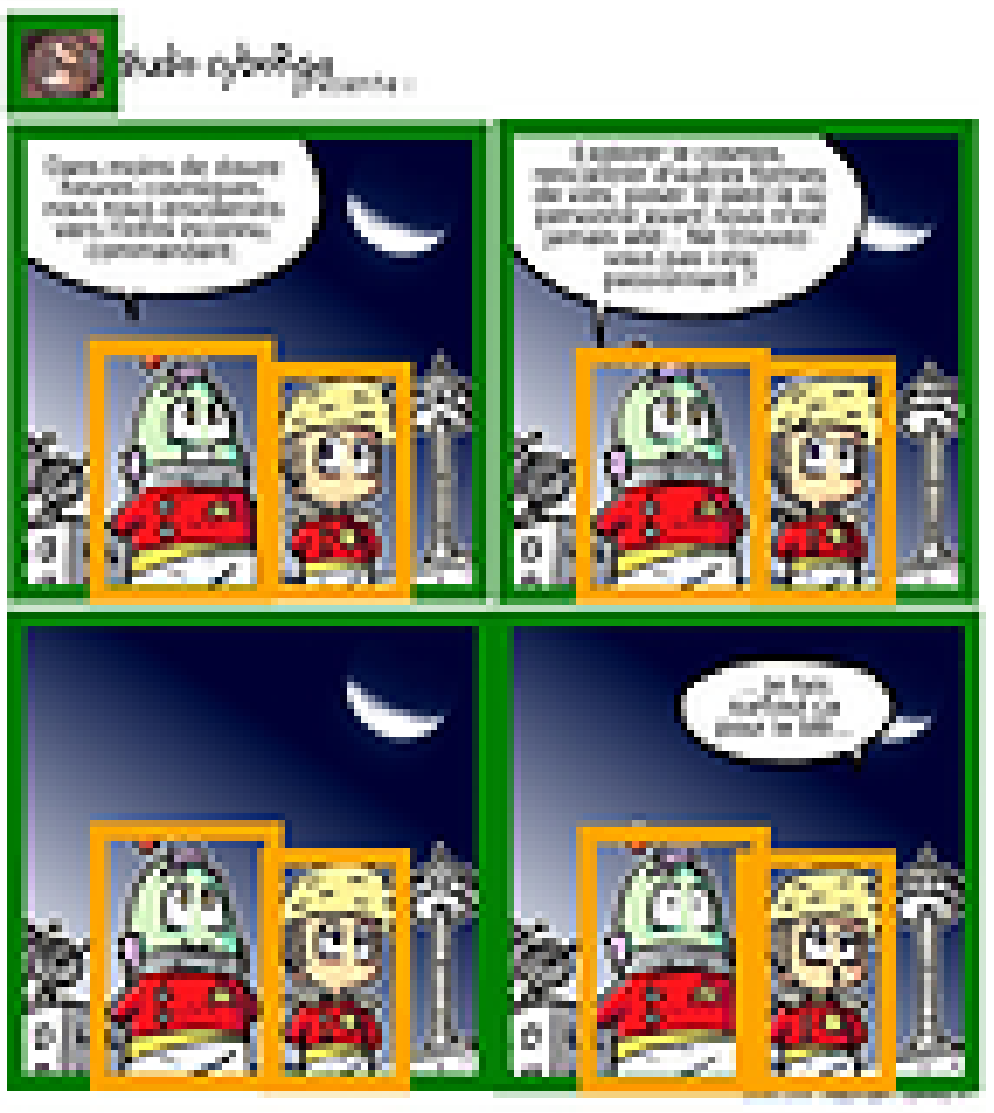}} &
    \frame{\includegraphics[width=0.25\hsize]{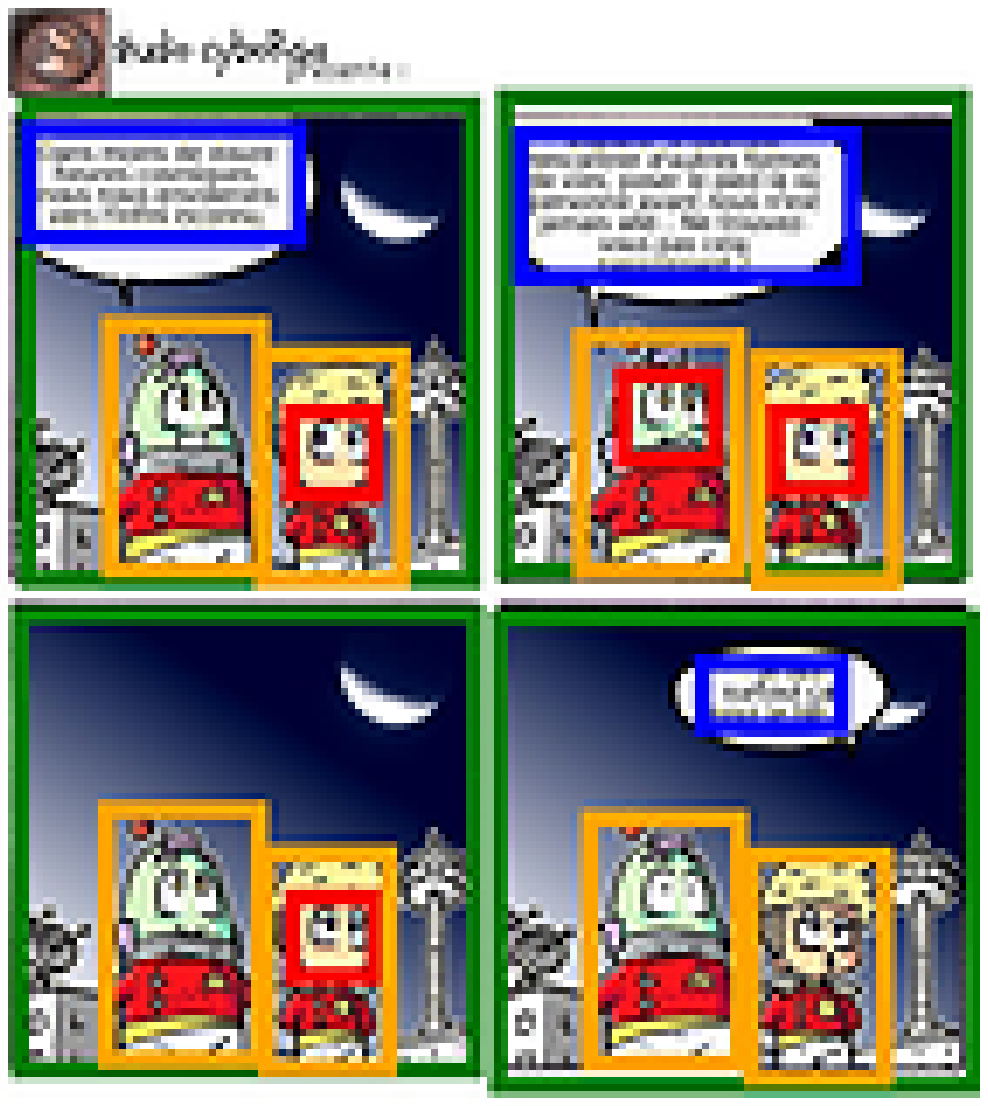}} \\

    Ground truth&SSD300-fork
  \end{tabular}
  \caption{Detection results of eBDtheque.
    Each annotation and detected object is represented by a rectangle (\textit{frame} (green), \textit{text} (blue), \textit{face} (red), and \textit{body} (orange)).
    We show \textit{frame} (green) and \textit{body} (orange) in the left column because eBDtheque does not provide annotations for \textit{text} and \textit{face}.}
  \label{fig:experiment:ebdtheque}
\end{figure}

To compare with existing comic object detection methods,
we applied SSD300-fork to eBDtheque~\cite{eBDtheque2013}.
Although eBDtheque mainly consists of French comics that have different drawing style compared to the comics in Manga109,
we did not tune our model for eBDtheque (the weights were trained using Manga109-annotations).
\Cref{fig:experiment:ebdtheque} shows some results on eBDtheque.
We also show the ground truth of \textit{frame} and \textit{body} in the left column.

\Cref{table:experiment:ebdtheque} shows the scores of existing methods (copied from \cite{rigaud2015knowledge}) and SSD300-fork.
We used the recall (R), the precision (P), and the F-measure (F) as metrics and
chose the threshold that maximized the F-measure.
For \textit{frame}, the existing method~\cite{rigaud2015knowledge} worked better.
This would be due to the differences in drawing styles between French comics and Japanese.
For example, some frames in eBDtheque are bent, where most frames in Japanese comics consist of straight lines.
SSD300-fork trained using Manga109-annotations might be overfitted to such Japanese-style frames.
Note that, although we did not use the images in eBDtheque for training, our SSD300-fork still achieved more than 70~\%.
For \textit{body}, SSD300-fork outperformed the existing method by a large margin (around 20~\% for the all measurement criteria).
This impressive result might be due to the fact that SSD300-fork was trained using diverse kinds of body images.
\begin{table}[!t]
  \centering
  \caption{The comparison with methods designed for comics using eBDtheque.
  For each category, three metrics are computed: Recall (R), Precision (P), and F-measure (F).}
  \label{table:experiment:ebdtheque}
  {\tabcolsep=1.5mm \begin{tabular}{@{}lrrrrrr@{}}
    \toprule
    &\multicolumn{3}{c}{frame}&\multicolumn{3}{c}{body} \\ \cmidrule(l){2-7}
    Method&R&P&F&R&P&F \\ \midrule
    Arai et al.~\cite{arai2011method}&58.0&75.3&65.6&\dash&\dash&\dash \\
    Rigaud et al.~\cite{rigaud2013robust}&78.0&73.2&75.5&\dash&\dash&\dash \\
    Rigaud et al.~\cite{rigaud2015knowledge}&\bf{81.2}&\bf{86.6}&\bf{83.8}&21.6&40.5&28.2 \\ \midrule
    SSD300-fork&73.3&76.4&74.8&\bf{42.2}&\bf{58.0}&\bf{48.8} \\
    \bottomrule
  \end{tabular}}
\end{table}

\footnotetext[10]{``UnbalanceTokyo'' \textcopyright Uchida Minako}

\section{Conclusion}
In this paper, we focused on the object detection task for comics.
There are two problems in applying the CNN-based detection methods to comic object detection:
the lack of a large-scale dataset and the assignment problem.

To solve the lack of a dataset, we created Manga109-annotations, which is a large-scale annotation dataset.
We manually annotated Manga109, which is an existing comics image dataset, with the help of 82 workers in eight months.
This annotation process contains double checking and refinement steps to improve the quality of annotations.
Finally, we obtained 527,685 annotations over 10,130 (double-sided) pages.
We annotated not only bounding boxes, required for object detection, but also character names and contents of texts.
These annotations can be used in future research.

For the assignment problem, we proposed SSD300-fork.
This model replicates the detection layer for each category to avoid the assignment problem.
We compared SSD300-fork with other CNN-based methods using Manga109-annotations,
and confirmed SSD300-fork achieved the best performance.
It outperformed SSD300, which is the base model of SSD300-fork, by 3~\% based on the mAP score.
In particular, for \textit{face} detection, the AP score increased by 9~\% against SSD300.

We also compared our method with detection methods that were designed for comics.
As these methods were evaluated using eBDtheque,
we applied SSD300-fork trained using Manga109-annotations to eBDtheque.
For \textit{frame}, our model archived 75~\% at the F-measure,
where the state-of-the-art method~\cite{rigaud2015knowledge} achieved 86~\%.
On the other hand, the score of \textit{body} was higher than the existing method by around 20~\%.

\bibliographystyle{spmpsci}      
\bibliography{references}

\end{document}